# Structure-Aware Stochastic Control for Transmission Scheduling

Fangwen Fu and Mihaela van der Schaar

Department of Electrical Engineering, University of California, Los Angeles, Los Angeles, CA, 90095

{fwfu, mihaela}@ee.ucla.edu

**ABSTRACT**

In this paper, we consider the problem of real-time transmission scheduling over time-varying channels. We first formulate the transmission scheduling problem as a Markov decision process (MDP) and systematically unravel the structural properties (e.g. concavity in the state-value function and monotonicity in the optimal scheduling policy) exhibited by the optimal solutions. We then propose an online learning algorithm which preserves these structural properties and achieves $\varepsilon$-optimal solutions for an arbitrarily small $\varepsilon$. The advantages of the proposed online method are that: (i) it does not require a priori knowledge of the traffic arrival and channel statistics and (ii) it adaptively approximates the state-value functions using piece-wise linear functions and has low storage and computation complexity. We also extend the proposed low-complexity online learning solution to the prioritized data transmission. The simulation results demonstrate that the proposed method achieves significantly better utility (or delay)-energy trade-offs when comparing to existing state-of-art online optimization methods.

## I. INTRODUCTION

Wireless systems often operate in dynamic environments where they experience time-varying channel conditions (e.g. fading channel) and dynamic traffic arrivals. To improve the energy efficiency of such systems while meeting the delay requirements of the supported applications, the scheduling decisions (i.e. determining how much data should be transmitted at each time) should be adapted to the time-varying environment [1][9]. In other words, it is essential to design scheduling policies which consider the time-varying characteristics of the channels as well as that of the applications (e.g. backlog in the transmission buffer, priorities of traffic, etc.). In this paper, we use optimal stochastic control to determine the transmission scheduling policy that maximizes the application utility given energy constraints.

The problem of energy-efficient scheduling for transmission over wireless channels has been intensively investigated in [1]-[15]. In [1], the trade-off between the average delay and the average energy consumption for a fading channel is characterized. The optimal energy consumption in the asymptotic large delay region (which corresponds to the case where the optimal energy consumption is close to the optimal energy consumption under queue stability constraints, as shown in Figure 1) is analyzed. In [8], joint source-channel coding is considered to improve the delay-energy trade-off. The structural properties of the solutions which achieve the optimal energy-delay trade-off are provided in [5][6][7]. It is proven that the optimal amount of data to be transmission increases as the backlog (i.e. buffer occupancy) increases, and decreases as the channel conditions degrade. It is also proven that the optimal state-value function (representing the optimal long-term utility starting from one state) is concave in terms of the instantaneous backlog.

Energy-efficient scheduling for traffic with individual delay deadlines is considered in [2][3][4]. In [2], the optimal scheduling policy is obtained using dynamic programming. In [3][4], optimality conditions are characterized for the optimal scheduling policies, and based on these, online heuristic scheduling policies are developed. Besides considering the time-varying channel conditions, the heterogeneous traffic features (e.g. different delay deadlines, importance and dependencies of packets) are considered in [14][15], where the optimal scheduling policies are developed by explicitly considering the impact of the heterogeneous data traffic.

We notice that the above solutions are characterized by assuming that the statistical knowledge of the underlying dynamics (e.g. channel state distribution, packet arrival distribution, etc.) is known. When the knowledge is unavailable, only heuristic solutions are provided, which cannot guarantee the optimal performance. In order to cope with the unknown environment, the stability-constrained optimization methods are developed in [10]-[13], where, instead of minimizing the queue delay, the queue stability is considered. The optimal energy consumption is achieved only for asymptotically large queue sizes (corresponding to asymptotic large delays, i.e. in large delay region). These methods do not provide optimal energy consumption in the small delay region which is shown in Figure 1.

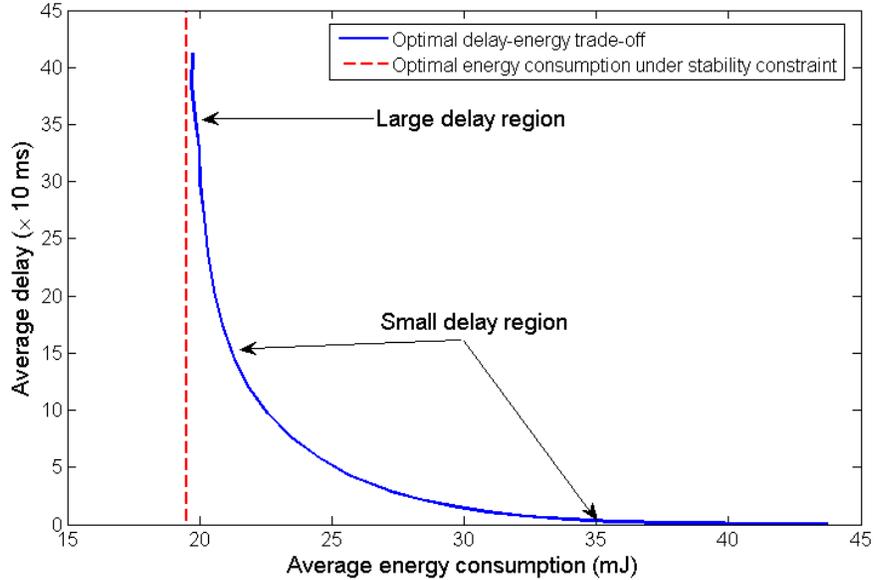

Figure 1. Illustration of large delay region and small delay region

Other methods for coping with transmission in unknown environment rely on online learning algorithms developed based on reinforcement learning for Markov decision processes in which the state-value function is learned online, at transmission time [16][17]. It has been proven that the online learning algorithms converge to optimal solutions when all the possible states are visited infinitely often. However, these methods have to learn the state-value function for each possible state and hence, they require large memory to store the state-value function (i.e. exhibit large memory overhead) and they take a long time to learn (i.e. exhibit a slow converge rate), especially when the state space is large, as in the considered wireless transmission problem.

In this paper, we consider a model similar to that of [1] with a single transmitter and a single receiver on a point-to-point wireless link where the system is time-slotted and the underlying channel state can be modelled as a finite-state Markov chain [20]. We first formulate the energy-efficient transmission scheduling problem as a constrained MDP problem. We then present the structural properties associated with the optimal solutions. Specifically, we show that the optimal state-value function is concave in terms of the backlog. Different from the proofs given in [6][7], we introduce a post-decision state (which is a "middle" state, in which the transmitter finds itself after a packet transmission but before the new packets' arrivals and new channel realization) and post-decision state-value function which provide an easier way to derive the structural results and build connections between the Markov decision process formulation and the queue stability-constrained optimization formulation. In this paper, we show that the stability-constrained optimization formulation is a special case in which the post-decision state-value function has a fixed form computed only based on the backlog and without considering the impacts of time-correlation of channel states.

In order to cope with the unknown time-varying environment, we develop a low-complexity online learning algorithm. Similar to the reinforcement learning algorithm [23], we update the state-value function online, when transmitting the data. However, different from the previous online learning algorithms [16][17], we approximate the state-value function using piece-wise linear functions, which allow us to represent the state-value function in a compact way, while preserving the concavity of the state-value functions. Instead of learning the state-value for each possible state, we only need to update the state-value in a limited number of states when using piece-wise linear approximation, which can significantly accelerate the convergence rate. We further prove that this online learning algorithm can converge to the $\varepsilon$-optimal [1] solutions, where $\varepsilon$ is controlled by a user defined approximation error tolerance. Our proposed method provides a systematic methodology for trading-off the complexity of the optimal controller and the achievable performance. As mentioned before, the stability-constrained optimization only uses the fixed post-decision state-value function (only considering the impacts of backlog), which can achieve the optimal energy consumption in the asymptotic large delay region, but often exhibits poor performance in the small delay region as shown in Section VI. However, our proposed method is able to achieve $\varepsilon$-optimal performance in both regions.

In order to consider the heterogeneity of the data, we further extend the proposed online learning algorithm to a more complicated scenario, where the data are prioritized and buffered into multiple priority queues. In general, the post-decision state-value function is multi-dimensional and needs to be learned online, which often requires high storage and computation complexity [26][27]. In contrast, using the priority queues, we are able to decompose the multi-dimensional post-decision state-value function into multiple single-dimensional concave post-decision state-value functions which enable us to learn them online, using our proposed structure-aware learning algorithm, which has low complexity and fast convergence rate.

---

[1] $\varepsilon$-optimal solutions mean that the solutions is within the $\varepsilon$-neighbourhood of the optimal solutions.

The difference between our proposed method and the representative methods presented in the literature is summarized in Table 1.

Table 1. Comparison between our proposed methods and other representative online optimization methods

|  | Statistical knowledge of unknown dynamics | Exploring structural properties | Performance | | Convergence rate | Storage complexity | Computation complexity |
| --- | --- | --- | --- | --- | --- | --- | --- |
|  |  |  | Large delay region | Small delay region |  |  |  |
| Stability-constrained optimization [10][11][12][13] | No | No | Asymptotically optimal | Suboptimal | No learning | Low | Low |
| Q-learning [17] | No | No | Convergence to optimal solutions | | Slow | Large | Low |
| Q-learning [16] | No | Yes | Convergence to optimal solutions | | Slow | Large | Low |
| Online learning with adaptive approximation (proposed) | No | Yes | Convergence to $\varepsilon$-optimal solutions | | Fast | Low | Low |

The paper is organized as follows. Section II formulates the transmission scheduling problem as a constrained MDP problem and presents the methods to solve it when the underlying dynamics are known. Section III introduces the concepts of post-decision state and post-decision state-value function for the considered problem. Section IV presented an approximate online learning for solving the MDP problem by exploring the structural properties of the solutions. Section V extends the online learning algorithms to the scenarios where the incoming traffic is heterogeneous (i.e. has different priorities). Section VI presents the simulation results, which is followed by the conclusions in Section VII.

## II. FORMULATING TRANSMISSION SCHEDULING AS CONSTRAINED MDP

In this paper, we consider a transmission scheduling problem in which one single user (a transmitter–receiver pair) transmits data from one finite transmission buffer (the data from multiple buffers are discussed in Section V) over a time-varying channel as shown in Figure 2. We assume a time-slotted transmission. The backlog at the transmitter side at time $t \in \mathbb{Z}_+$ is denoted by $x_t \in [0, B]$, where $B$ is the capacity of the buffer. At time $t$, the user transmits the amount of $y_t \in [0, x_t]$ data. The traffic arrival takes place at the end of each time slot. The traffic arrival at time $t$ is denoted by $a_t \in \mathbb{R}_+$. For simplicity[2], we assume that the traffic arrival $a_t$ is an i.i.d. random variable, which is independent of the channel conditions and buffer sizes [1]. We further assume that the channel conditions are constant within one time slot but that they vary across time slots. The channel state at time $t$ is denoted by $h_t \in \mathcal{H}$, where $\mathcal{H}$ is the finite set of possible channel conditions. The channel state transition across the time slots is modelled as a finite state Markov chain [20] and the transition probability is denoted by $p_h(h_{t+1} \mid h_t)$, which is independent of the buffer size and the traffic arrival.

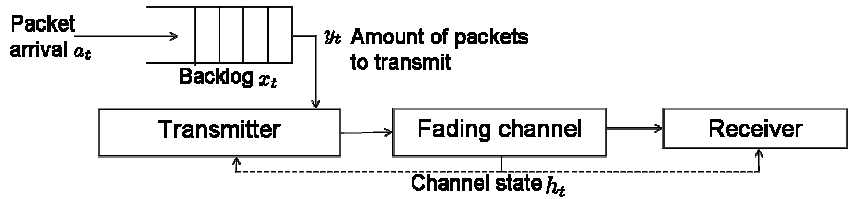

Figure 2.    Transmission scheduling model for a single user

At each time $t$, the buffer dynamics are captured by the following expression:

$$x_{t+1} = \min(x_t - y_t + a_t, B) \tag{1}$$

When the amount of $y_t$ data is transmitted, the received immediate utility by the user at time $t$ is $u(x_t, y_t) \geq 0$ and the incurred transmission cost is $c(h_t, y_t) \geq 0$. The immediate utility can be the negative value of the backlog when minimizing the delay as considered in the simulation in Section VI. The transmission cost can be the consumed energy. In this paper, we assume that the utility function and the transmission cost function are known a-priori and satisfy the following conditions.

*Assumption 1*: $u(x, y)$ is supermodular and jointly concave in $(x, y)$;

---

[2] The method proposed in this paper can be easily extended to the case in which the packet arrival is Markovian by defining an extra arrival state [8].

*Assumption 2*: $c(h,y)$ is increasing and convex in $y$ for any given $h \in \mathcal{H}$.

The supermodularity is defined as follows.

**Definition**: A function $f: X \times Y \to \mathbb{R}$ is a supermodular in the pair of $(x,y)$, if for all $y' \geq y, x' \geq x$,

$$f(x',y') - f(x',y) \geq f(x,y') - f(x,y). \tag{2}$$

We note that the assumption of supermodularity and joint concavity on the utility functions is reasonable and has been widely used in the past work [16]. The main reason for introducing the supermodular concept is to establish the monotonic structure of the optimal scheduling policy. That is, the optimal scheduling policy $\pi(x)$ given by

$$\pi(x) = \arg\max_{y \in Y} f(x,y) \tag{3}$$

is non-decreasing in $x$. The property is established by Topkis [22]. By assuming that the utility function $u(x,y)$ is supermodular, we will prove that the optimal scheduling policy for the considered problem will also satisfy the monotonic structure.

The increasing assumption on the transmission cost $c(h,y)$ represents the fact that transmitting more data results in higher transmission cost at the given channel condition $h$. We introduce the convexity on the transmission cost in order to capture the self-congestion effect [6] of the data transmission.

The objective for the user is to maximize the long-term utility under the constraint on the long-term transmission cost:

$$\begin{aligned} \max_{y_t \geq 0, \forall t} & \; E\left[\sum_{t=0}^{\infty} \alpha^t u(x_t, y_t)\right] \\ \text{s.t. } & \; E\left[\sum_{t=0}^{\infty} \alpha^t c(h_t, y_t)\right] \leq \overline{c} \end{aligned} \tag{4}$$

where $\alpha$ is the discount factor in the range of $[0,1)$ and $\overline{c}$ is the budget on the transmission cost. In this formulation, the long-term utility (transmission cost) is defined as the discounted sum of utility (transmission cost). When $\alpha \to 1$, the optimal solution to the optimization in (4) is equivalent to the optimal solution to the problem maximizing the average utility under the average transmission cost constraint as considered in [1].

The optimization in (4) can be formulated as a constrained Markov decision process. We define the state at time $t$ as $s_t = (x_t, h_t) \in [0, B] \times \mathcal{H}$ and the action at time $t$ is $y_t$. Then, the scheduling control is a Markovian system with the state transition probability:

$$p(s_{t+1} \mid s_t, y_t) = p(a_t) p(h_{t+1} \mid h_t) \delta(x_{t+1} - \min(x_t - y_t + a_t, B)) \tag{5}$$

where $\delta(z)$ is a Kronecker delta function, i.e. $\delta(z) = 1$ if $z = 0$ and $\delta(z) = 0$ otherwise.

For this constrained MDP problem, we define the scheduling policy as a function mapping the current state $s_t$ to the current action $y_t$ and denote it by $\pi(\cdot)$. The set of possible policies is denoted by $\Phi$. The long-term utility and transmission cost associated with the policy $\pi$ are denoted by $U^\pi(s_0)$ and $C^\pi(s_0)$, and can be computed as:

$$U^\pi(s_0) = E\left[\sum_{t=0}^{\infty} \alpha^t u(x_t, \pi(s_t)) \mid s_0\right], \tag{6}$$

and

$$C^\pi(s_0) = E\left[\sum_{t=0}^{\infty} \alpha^t c(h_t, \pi(s_t)) \mid s_0\right]. \tag{7}$$

Any policy $\pi^*$ that maximizes the long-term utility under the transmission cost constraint is referred to as the optimal policy. The optimal utility associated with the optimal policy is denoted by $U^*_{\bar{c}}(s_0)$, where the subscript indicates that the optimal utility depends on $\bar{c}$. By introducing the Lagrangian multiplier associated with the transmission cost, we are able to transform the constrained MDP into an unconstrained MDP problem. From [18], we know that solving the constrained MDP problem is equivalent to solving the unconstrained MDP and its Lagrangian dual problem. We present this result in Theorem 1 without proof. The detailed proof can be founded in Chapter 6 of [18].

**Theorem 1**: The optimal utility of the constrained MDP problem can be computed by

$$U^*_{\bar{c}}(s_0) = \max_{\pi \in \Phi} \min_{\lambda \geq 0} J^{\pi,\lambda}(s_0) + \lambda \bar{c} = \min_{\lambda \geq 0} \max_{\pi \in \Phi} J^{\pi,\lambda}(s_0) + \lambda \bar{c}, \tag{8}$$

where

$$J^{\pi,\lambda}(s_0) = E\left[\sum_{t=0}^{\infty} \alpha^t (u(x_t, \pi(s_t)) - \lambda c(h_t, \pi(s_t))) \mid s_0\right], \tag{9}$$

and a policy $\pi^*$ is optimal for the constrained MDP if and only if

$$U^*_{\bar{c}}(s_0) = \min_{\lambda \geq 0} J^{\pi^*,\lambda}(s_0) + \lambda \bar{c}. \tag{10}$$

We note that the maximization in the rightmost expression in Eq. (8) can be performed as an unconstrained MDP given the Lagrangian multiplier. Solving the unconstrained MDP is equivalent to solving the Bellman's equations which is presented in the following:

$$J^{*,\lambda}(s) = \max_{\pi \in \Phi}\left[u(x,\pi(s)) - \lambda c(h,\pi(s)) + \alpha \sum_{s' \in S} p(s' \mid s, \pi(s)) J^{*,\lambda}(s')\right], \forall s \tag{11}$$

We will discuss how to solve the Bellman's equation in Eq. (11). We denote the optimal scheduling policy associated with the Lagrangian multiplier $\lambda$ as $\pi^{*,\lambda}$. The long-term transmission cost associated with the scheduling policy $\pi^{*,\lambda}$ is given by

$$C^{\pi^{*,\lambda}}(s_0) = E\left[\sum_{t=0}^{\infty} \alpha^t c\left(h_t, \pi^{\lambda,*}(s_t)\right) \mid s_0\right] \tag{12}$$

It was proved in [18] that the long-term transmission cost $C^{\pi^{*,\lambda}}(s_0)$ is a piece-wise linear non-increasing convex function of the Lagrangian multiplier $\lambda$. Then, a simple algorithm to find the optimal Lagragian multiplier $\lambda^*$ can be found through the following update:

$$\lambda_{n+1} = \max\left(\lambda_n + \gamma_n \left(C^{\pi^{*,\lambda_n}}(s_0) - \bar{c}\right), 0\right) \tag{13}$$

where $\gamma_n = \frac{1}{n}$. The convergence to the optimal $\lambda^*$ is ensured due to the fact that $C^{\pi^{*,\lambda}}(s_0)$ is a piece-wise convex function of Lagrangian multiplier $\lambda$.

## III. POST-DECISION STATE BASED DYNAMIC PROGRAMMING

In this section and subsequent sections, we will discuss how to solve the Bellman's equations in Eq. (11) by exploring the structural properties of the optimal solution for our considered problem. From Eq. (11), we note that the expectation (over the data arrival and channel transition) is embedded into the term to be maximized. However, in a real system, the distribution of the data arrival and channel transition is often unavailable a priori, which makes it computationally impossible to compute the expectation exactly. It is possible to approximate the expectation using sampling, but this significantly complicates the maximization. Similar to [23][28], we introduce an intermediate state which represents the state after scheduling the data but before the new data arrives and new channel state is realized. This intermediate state is referred to as the post-decision state $\tilde{s} = (\tilde{x}, \tilde{h})$. In order to differentiate the "post-decision" state $\tilde{s}_t$ from the state $s_t$, we refer to the state $s_t$ as the "normal" state. The post-decision state at time slot $t$ is also illustrated in Figure 3. From this figure, we know that the post-decision state is a deterministic function of the normal state $s_t$ and the decision $y_t$ which is given by:

$$\tilde{x}_t = x_t - y_t, \quad \tilde{h}_t = h_t \tag{14}$$

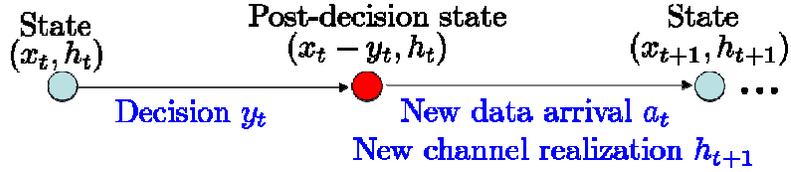

Figure 3. Illustration of post-decision state

By defining the value function $V^{*,\lambda}(x,h)$ for the post-decision state $\tilde{s} = (x,h)$, the Bellman's equations in Eq. (11) can be rewritten as follows:

$$V^{*,\lambda}(x,h) = \sum_a \sum_{h' \in \mathcal{H}} p(a) p(h' \mid h) J^{*,\lambda}(\min(x+a,B), h') \tag{15}$$

$$J^{*,\lambda}(x,h) = \max_{0 \leq y \leq x} \left[ u(x,y) - \lambda c(h,y) + \alpha V^{*,\lambda}(x-y,h) \right] \tag{16}$$

The first equation shows that the post-decision state-value function $V^{*,\lambda}(\cdot,\cdot)$ is obtained from the normal state-value function $J^{*,\lambda}(\cdot,\cdot)$ by taking the expectation over the possible traffic arrivals and possible channel transitions. The second equation shows that the normal state-value function is obtained from the post-decision state-value function $V^{*,\lambda}(\cdot,\cdot)$ by performing the maximization over the possible scheduling actions. This maximization is referred to as the foresighted optimization since the optimal scheduling policy is obtained by maximizing the long-term utility.

The advantages of introducing the post-decision state and corresponding value functions are summarized next.

- In the normal state-based Bellman's equations in Eq. (11), the expectation over the possible channel states has

to be performed before the maximization over the possible scheduling actions. Hence, performing the maximization requires the knowledge of the data arrival and channel dynamics. In contrast, in the post-decision state-based Bellman's equations in Eqs. (15) and (16), the expectation over the possible arrivals and channel states is separated from the maximization. If we directly approximate the post-decision state-value function online (which will be detailed out in Section IV), we can perform the maximization without computing the expectation and hence, without the knowledge of the data arrival and channel dynamics.

- By introducing the post-decision states, the foresighted optimization in Eq. (16) is deterministic. We will further show that the post-decision state-value function for our considered problem is concave in the backlog $x$ in Section IV. Hence, the foresighted optimization in Eq. (16) is a one-variable convex optimization and can be easily solved using the large library of solvers (e.g. CVX [35]) for the deterministic convex problem.

- Since the post-decision state-value function is concave (shown in Section IV), we are able to compactly represent the post-decision state-value functions using piece-wise linear function approximations which preserve the concavity and the structure of the problem.

- As depicted in Figure 2, we notice that the channel and traffic dynamics are independent of the queue length[3], which enables us to develop a batch update on the post-decision state-value function described in Section IV.

The Bellman's equations for the scheduling problem can be solved using value iteration, policy iteration or linear programming, etc., when the dynamics of channel and traffic are known a-priori. However, in an actual transmission system, this information is often unknown a-priori. In this case, instead of directly solving the Bellman's equations, online learning algorithms have been developed to update the state-value functions in real time, e.g. Q-learning [16][17], actor-critic learning [23], etc. However, these online learning algorithms often experience slow convergence rates. In this paper, we develop a low-complexity online learning algorithm which can significantly increase the convergence rate.

## IV. APPROXIMATE DYNAMIC PROGRAMMING

In this section, we will first develop the structural properties of the optimal scheduling policy and corresponding post-decision state-value function, based on which we will then discuss the approximation of the post-decision state-value function and the online learning of the post-decision state-value function. This approximation allows us to compactly represent the post-decision state-value function. The following theorem shows that the optimal post-decision state-value function $V^{*,\lambda}(x,h)$ is concave in the parameter $x$.

**Theorem** 2. With assumptions 1 and 2, the post-decision state-value function $V^{*,\lambda}(x,h)$ is a concave function in $x$ for any given $h \in \mathcal{H}$ and the optimal scheduling policy $\pi^{*,\lambda}(x,h)$ is non-decreasing in $x$ for any given $h \in \mathcal{H}$.

Proof: See Appendix A. The key idea of proving this theorem is also illustrated in Figure 4 where the concavity of $V^{*,\lambda}(x,h)$ and the non-decreasing property of $\pi^{*,\lambda}(x,h)$ is proved using backward induction.

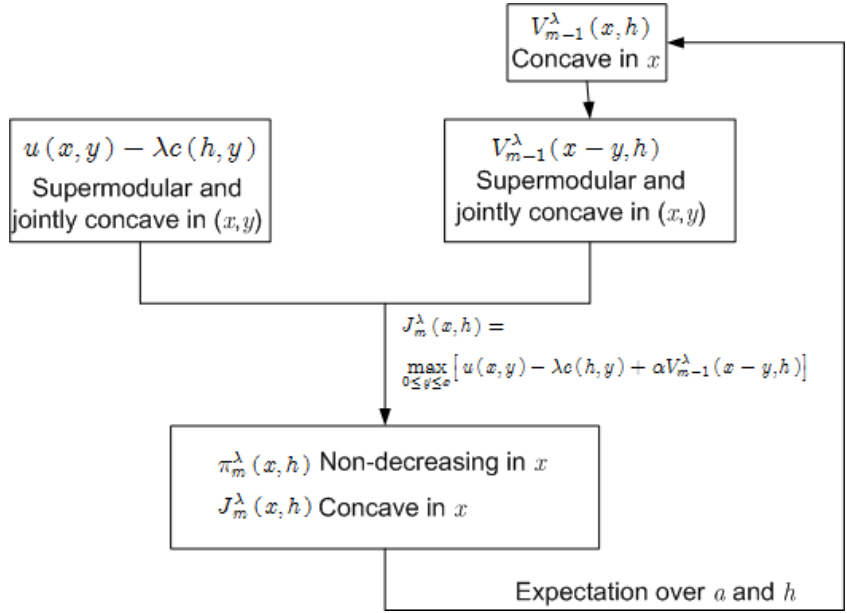

Figure 4. Key idea in proving Theorem 2.

In the above, we derive the structural properties associated with the optimal solutions. However, we still face two problems: (i) since the queue length is often continuous or the capacity of the queue is large, the state space is very large, thereby leading to expensive computation cost and storage overheads; (ii) the channel states and incoming traffic data dynamics are often difficult to characterize a priori, such that the Bellman's equation cannot be solved before the actual traffic transmission. In this section, we first present an approximation method to compactly represent the post-decision state-value function. We then present approximate dynamic programming solutions using the approximated state-value function. To deal with the unknown dynamics, we propose an online learning algorithm based on the approximated state-value function.

---

[3] If the channel and traffic dynamics depends on the queue length, we can still separate the maximization and expectation. However, the update on the post-decision state-value function is much more complicated and will be investigated in the future.

A.  **Approximating the post-decision state-value function**

In this section, we present the proposed method for approximating the post-decision state-value function and we quantify the gap between the approximated post-decision state-value function and optimal post-decision state-value function. We define the post-decision state based dynamic programming operator as

$$TV^\lambda(x,h) = \sum_a \sum_{h' \in \mathcal{H}} p(a) p(h' \mid h)$$
$$\max_{0 \leq y \leq \min(x+a,B)} \left[ u(\min(x+a,B), y) - \lambda c(h', y) + \alpha V^\lambda(\min(\tilde{x}+a,B) - y, h') \right] \tag{17}$$

It can be proved that the operator $T$ is a maximum norm $\alpha$-contraction [19], i.e. $\left\| TV^\lambda - TV'^\lambda \right\|_\infty \leq \alpha \left\| V^\lambda - V'^\lambda \right\|_\infty$ and $\lim_{t \to \infty} T^t V^\lambda = V^{*,\lambda}$ for any $V^\lambda$. Due to the concavity preservation of the post-decision state based dynamic programming operator, we choose the initial post-decision state-value function as a concave function in the queue length $x$ and denoted as $V_0^\lambda$.

In Appendix B, we present a method to approximate a concave function using piece-wise linear function. In this method, we are able to control the computation complexity and achievable performance by using different predetermined approximation error threshold $\delta$. The advantage of the proposed approximation method is that we can approximate the concave function only by evaluating the function at a limited number of points and without knowing the closed form of the function. We denote the approximation operator developed in Appendix B as $A_\delta f$ for any concave function $f$. Then $A_\delta f$ is a piece-wise linear concave function and satisfies $0 \leq f - A_\delta f \leq \delta$. The dynamic programming operator with adaptive approximation is represented by $A_\delta TV^\lambda$.

**Theorem 3**: Given the initial piece-wise linear concave function $V_0^\lambda$, we have

(i). 
$$0 \leq V^{*,\lambda} - (A_\delta T)^\infty V_0^\lambda \leq \frac{\delta}{1-\alpha} \qquad (18)$$

(ii). For any other concave function $V_0'^{\lambda}$,

$$\left\| (A_\delta T)^\infty V_0^\lambda - (A_\delta T)^\infty V_0'^{\lambda} \right\|_\infty \leq \frac{\delta}{1-\alpha} \qquad (19)$$

Proof: See Appendix.

In Theorem 3, we notice that, starting with any piece-wise linear concave function $V_0^\lambda$, the value iteration using the dynamic programming operator with adaptive approximation converges to the $\varepsilon$-optimal post-decision state-value function $V^{*,\lambda}$, where $\varepsilon = 1/(1-\alpha)$. For any arbitrarily small $\varepsilon$, we can choose $\delta = \varepsilon(1-\alpha)$. We further notice that, in Theorem 3, the convergence is achieved by applying the dynamic programming operator, which requires the statistical knowledge of the underlying dynamics. In the next section, we present how to learn the post-decision state-value function online with this adaptive approximation, which does not require this statistical knowledge.

### B. Online learning by approximating the post-decision state-value function

In section IV.A, we propose an approximated dynamic programming to compute the post-decision state-value function assuming that the traffic and channel dynamics are known a-priori. However, the traffic and channel dynamics cannot be often characterized beforehand. When the dynamics of the channel and data arrivals are not known before the transmission system is implemented, we face the following difficulties: (i) the Bellman's equations for both the normal states and post-decision states cannot be explicitly solved since they require the distribution of traffic arrivals and the probability of channel transition; (ii) solving the Bellman's equation often requires multiple iterations to reach the optimal policy, e.g. using value iteration or policy iteration. Instead, in this section, we propose an online learning algorithm which estimates the post-decision state-value function (represented by the piece-wise linear function) online.

The Bellman's equations provide us the necessary foundations and principles to learn the optimal state-value functions and optimal policy on-line. From the observation presented in Section III, we note that the expectation is separated from the maximization when the post-decision state is introduced. We note that the online learning algorithm proposed in [17] only updates the post-decision state-value function in one post-decision state $(x_{t-1}, h_{t-1})$ in one time slot, which is referred to the one-state-per-time-slot online learning. However, in our considered transmission system, we notice that the data arrival probabilities and channel state transition are independent of the backlog $x$. In the other words, at time slot $t$, the traffic arrival $a_{t-1}$ and new channel state $h_t$ can be realized at any possible backlog $x$. Hence, instead of updating the post-decision state-value function only at the state $(\tilde{x}_{t-1}, h_{t-1})$, we are able to update the post-decision state-value function at all the states which have the same channel state $h_{t-1}$. From Eq. (15), we note that, given the traffic arrival $a_{t-1}$, new channel state $h_t$, and the post-decision state-value function $V^{t-1,\lambda}(x,h)$, we can obtain the optimal scheduling by solving the foresighted optimization:

$$J^{t,\lambda}(x, h_t) = \max_{0 \leq y \leq x} \left[ u(x,y) - \lambda c(h_t, y) + \alpha V^{t-1,\lambda}(x-y, h_t) \right] \qquad (20)$$

where $x = \min(\tilde{x} + a_{t-1}, B)$ and $\tilde{x}$ is the post-decision backlog.

As we point out at the beginning of this section, the statistics of the traffic arrival and channel state transition is not available beforehand. In this case, instead of computing the post-decision state-value function as in Eq. (15), we can update online the post-decision state-value function using reinforcement learning [23]. Specifically, at time slot $t-1$, the post-decision state is $(\tilde{x}_{t-1}, h_{t-1})$. At time slot $t$, the normal state becomes $s_t = (x_t, h_t)$ with $x_t = \min(\tilde{x}_{t-1} + a_{t-1}, B)$ and the channel state $h_{t-1}$ transits to $h_t$ with the unknown probability $p_h(h_t \mid h_{t-1})$. We can find the optimal scheduling policy at *any* normal state $s = (x, h_t)$, where $x = \min(\tilde{x} + a_t, B), \forall \tilde{x}$, by solving the optimization in Eq. (20) which gives the normal state-value function $J^{t,\lambda}(x, h_t)$. From Eq. (15), we note that the post-decision state-value function is computed by taking the expectation of the normal state-value function over all possible traffic arrival and channel transmissions. However, instead of taking expectation, we can update the post-decision state-value function using time-averages for all the states $\{(\tilde{x}, h_{t-1}), \forall \tilde{x}\}$ as follows:

$$\begin{aligned} V^{t,\lambda}(\tilde{x}, h_{t-1}) &= (1-\beta_t) V^{t-1,\lambda}(\tilde{x}, h_{t-1}) + \beta_t J^{t,\lambda}(\min(\tilde{x}+a_t, B), h_t), \forall \tilde{x} \\ V^{t,\lambda}(\tilde{x}, \tilde{h}) &= V^{t-1,\lambda}(\tilde{x}, \tilde{h}), \forall \tilde{x}, \tilde{h} \neq \tilde{h}_{t-1} \end{aligned} \qquad (21)$$

where $\beta_t$ is a learning rate factor [23], e.g. $\beta_t = 1/t$. We refer to the update as the "batch update" since it can update the post-decision state-value function $V(\tilde{x}, h_t)$ at all the states $\{(\tilde{x}, h_{t-1}), \forall \tilde{x}\}$.

The following theorem shows that the above online learning algorithm converges to the optimal post-decision state-value function $V^{*,\lambda}(x,h)$ based on which the optimal scheduling policy can be determined.

**Theorem 4**. The online learning on the post-decision state-value function $V^{t,\lambda}(x,h)$ converges to the optimal $V^{*,\lambda}(x,h)$ when the learning rate factor $\beta_t$ satisfies[4] $\sum_{t=0}^{\infty} \beta_t = \infty, \sum_{t=0}^{\infty} \beta_t^2 < \infty$.

Proof: See Appendix D.

We notice that, unlike the traditional Q-learning algorithm where the state-value function is updated for one state per time slot, our proposed online learning algorithm is able to update the post-decision state-value function for all the states $\{(\tilde{x}, h_{t-1}), \forall \tilde{x}\}$ in one time slot. The downside of the proposed online learning algorithm is that it has to update the post-decision state-value function $V^{t,\lambda}(\tilde{x}, h_{t-1})$ for all the states of $\{(\tilde{x}, h_t), \forall \tilde{x}\}$, which often requires many computations when the number of queue states is large. To overcome this obstacle, we propose to approximate the post-decision state-value function $V^{t,\lambda}(\tilde{x}, h)$ using piece-wise linear functions since $V^{t,\lambda}(\tilde{x}, h)$ is a concave function. Consequently, instead of updating all the states for the post-decision state-value function, we only update a necessary number of states at each time slot, which is determined by our proposed adaptive approximation method presented in Appendix B. Specifically, given the traffic arrival $a_{t-1}$, new channel state $h_t$, and the approximated post-decision state-value function $\hat{V}^{t-1,\lambda}$ at time slot $t-1$, we can obtain the optimal scheduling $\pi(x, h_t)$ where $x = \min(\tilde{x} + a_{t-1}, B), \forall \tilde{x}$ and the state-value function $J^{t,\lambda}(x, h_t)$ by replacing the post-decision state-value function $V^{t-1,\lambda}$ in Eq. (20) with the approximated post-decision state-value function $\hat{V}^{t-1,\lambda}$. We can then update the post-decision state-value function $V^{t,\lambda}(\tilde{x}, h_{t-1})$ the same as in Eq. (21). However, as discussed in the above, we need to avoid updating the post-decision state-value function at all the states. It has been proved in Section IV.A, the post-decision state-value function $V^{t,\lambda}(\tilde{x}, h_{t-1})$ is a concave function. Hence, we propose to approximate the post-decision state-value function $V^{t,\lambda}(\tilde{x}, h_{t-1})$ by a piece-wise linear function which can preserves the concavity of the post-decision state-value function. The online learning algorithm is summarized in Algorithm 1.

Algorithm 1: Online learning algorithm with adaptive approximation

---
**Initialize**: $\hat{V}^{0,\lambda}(\cdot, h) = 0$ for all possible channel state $h \in \mathcal{H}$ ; post-decision state $s_0 = (x_0, h_0)$; $t = 1$.
**Repeat**:
  Observe the traffic arrival $a_{t-1}$ and new channel state $h_t$;
  Compute the normal state $(\min(x_{t-1} + a_{t-1}, B), h_t)$;
  Approximate the post-decision state-value function given by
    $\hat{V}^{t,\lambda}(x, h_{t-1}) = A_\delta \left( (1-\beta_t) \hat{V}^{t-1,\lambda}(x, h_{t-1}) + \beta_t J^{t,\lambda}(\min(x + a_t, B), h_t) \right)$;
  Compute the optimal scheduling policy $y^{t,*}$ and transmit the traffic;
  Update the post-decision state $s_t = \left( \min(x_{t-1} + a_{t-1}, B) - y^{t,*}, h_t \right)$;
  $t \leftarrow t + 1$;
**End**

---

The following theorem shows that the post-decision state-value function learned using Algorithm 2 converges to the $\varepsilon$-optimal post-decision state-value function.

**Theorem** 5: Given the concave function operator $A_\delta$ and the initial piece-wise linear concave function $V^{0,\lambda}(\cdot, h)$ for any possible channel state $h \in \mathcal{H}$, we have that

(i). $\hat{V}^{t,\lambda}(\cdot, h)$ is a piece-wise linear concave function;

(ii). $0 \leq V^{*,\lambda}(\cdot, h) - \hat{V}^{\infty,\lambda}(\cdot, h) \leq \delta/(1-\alpha)$ where $V^{*,\lambda}(\cdot, h)$ is the optimal post-decision state-value function.

Proof: See Appendix E.

Theorem 5 shows that, under the proposed online learning with adaptive approximation, the learned post-decision state-value function converges to the $\varepsilon$-optimal post-decision state-value function where $\varepsilon = \delta/(1-\alpha)$ and can be controlled by choosing different approximation error threshold $\delta$. In Section VI.A, we will show how the approximation error threshold affects the online learning performance.

It is worth to note that, the online learning algorithm with adaptive approximation shown in Algorithm 1 requires to be performed at each time slot, which may still have high computation complexity, especially when the number of states to be evaluated is large. In order to further reduce the computation complexity, we propose to update the post-decision state-value function (using the latest information about channel state transition and packet arrival) every $T$ $(1 \leq T < \infty)$ time slots. The following theorem shows that the online learning performed every $T$ time slots still converges to the $\varepsilon$-optimal solution.

---

[4] This conditions are quite normal and have been adopted in the literature [16][17].

**Theorem** 6: Given the concave function operator $A_\delta$ and the initial piece-wise linear concave function $V^{0,\lambda}(\cdot, h)$ for any possible channel state $h \in \mathcal{H}$, if the online learning algorithm shown in Algorithm 1 is performed every $T$ time slots, and the underlying channel state transition is an aperiodic Markov chain, then we have that

(i). $\hat{V}^{t,\lambda}(\cdot, h)$ is a piece-wise linear concave function;

(ii). $0 \leq V^{*,\lambda}(\cdot, h) - \hat{V}^{\infty,\lambda}(\cdot, h) \leq \delta/(1-\alpha)$.

Proof: The proof is the same as the one to Theorem 5. When the underlying channel state transition is aperiodic, updating the post-decision state-value function every $T$ time slots will still ensure that every state will be visited infinite times and hence will converge to the $\varepsilon$-optimal solution.

In Section VI.A, we also show the impact of choosing different $T$ s on delay-energy consumption trade-off.

## C. Comparison with other representative methods for single user transmission

In this section, we compare our online learning solution with the stability-constrained optimization proposed in [10][11][12][13] and the Q-learning algorithm proposed in [17] when applied to the single-user transmission. In the stability-constrained optimization, a Lyapunov function is defined for each state $(x_t, h_t)$ as $U(x_t, h_t) = x_t^2$. Note that the Lyapunov function only depends on the queue state $x_t$. Then, instead of minimizing the trade-off between the delay and the energy consumption, the stability-constrained optimization minimizes the trade-off between the Lyaponov drift (between the current state and post-decision state) and energy consumption:

$$\min_{0 \leq y \leq x} \lambda c(h_t, y_t) + x_t^2 - (x_t - y_t)^2 \qquad (22)$$

Compared to the foresighted optimization in Eq. (16), we note that, in the stability-constrained optimization method, the post-decision state-value function is approximated by[5]

$$V^\lambda (x_t - y_t, h_t) = \left(-(x_t - y_t) + (x_t - y_t)^2 - x_t^2\right)/\alpha.  \tag{23}$$

The difference between our proposed method and stability-constrained method can be summarized as follows:

(i). From Eq. (23), we note that the approximated post-decision state-value function is only a function of the current backlog $x_t$ and the scheduling decision $y_t$ and does not take into account the impact of the channel state transition and transmission cost. In contrast, we approximate the post-decision state-value function directly based on the optimal post-decision state-value function which explicitly considers the channel state transition and the transmission cost.

(ii). It has been proved [13] using the stability-constrained optimization method that the queue length must be larger than or equal to $\Omega(\sqrt{\lambda})$ [6], when the energy consumption is within $O(1/\lambda)$ [7] of the optimal energy consumption for the stability constraint, and that it asymptotically achieves the optimal trade-off between energy consumption and delay when $\lambda \to \infty$ (corresponding to the large-delay region). However, it provides poor performance in the small delay region (when $\lambda \to 0$ which allows us to have small queue length). This point is further examined in the numerical simulations presented in Section VI. In contrast, our proposed method is able to achieve the near-optimal solution in the small delay region.

(iii). Furthermore, in order to consider the average energy consumption constraint, a virtual queue has been maintained to update the trade-off parameter $\lambda$ in [13]. It can only be shown that this update achieves asymptotical optimality in the large delay region and results in very poor performance in the small delay region. Instead, we propose to update $\lambda$ using stochastic subgradients which achieves the $\varepsilon$-optimal solution in the small delay region, similar to [16][17].

---

[5] In [10][11][12][13], the utility function at each time slot is implicitly defined as $u(x_t, y_t) = -(x_t - y_t)$ representing the negative value of the post-decision backlog.

[6] $\Omega(\sqrt{\lambda})$ denotes that a function increases at least as fast as $\sqrt{\lambda}$.

[7] In [13], the parameter $V$ is used istead of $\lambda$.

We notice that the Q-learning[8] algorithm is also performed online. However, instead of updating the post-decision state-value function, it updates the state-action value function only at the visited state-action pair per time slot. The downsides of the Q-learning are: (i). it has to maintain a table to store the state-action value function for each state-action pair which is significantly larger than the state-value function table; (ii). It only updates the entry in the table at each time slot and does not preserve the structure of the considered problem. However, in our proposed online learning with adaptive approximation, we are able to approximate the post-decision state-value function using piece-wise linear function which requires to store the values of only a limited number of post-decision states. It also updates the post-decision state-value function at multiple states per time slot and further preserves the concavity of the post-decision state-value function. We show in the simulation results that our proposed online learning algorithm significantly accelerates the learning rate compared to the Q-learning algorithm.

In terms of computation complexity, we notice that the stability-constrained optimization performs the maximization shown in (22) once for the visited state at each time slot and the Q-learning algorithm also performs the maximization (finding the optimal state-value function from the state-action value function [23]) once for the visited state at each time slot. In our proposed online learning algorithm, we need to perform the foresighted optimization for the visited state at each time slot. Furthermore, we have to update the post-decision state value function at the evaluated states. The number of states to be evaluated at each time slot is denoted by $n_\delta$ which is determined by the approximation error threshold $\delta$. If we update the post-decision state-value function every $T$ time slots, then the total number of foresighted optimization to be performed, on average, is $1 + \frac{n_\delta}{T}$. From the simulation results, we notice that we can often choose $n_\delta < T$ which means that the number of foresighted optimization to be performed per time slot is less than 2.

## V. APPROXIMATE DYNAMIC PROGRAMMING FOR MULTIPLE PRIORITY QUEUES

In this section, we consider that the user delivers prioritized data, buffered in multiple queues. The backlog

---

[8] Q-learning algorithms may not require to know the utility functions. However, in our paper, the utility function is assumed to be known.

state is then denoted by $\boldsymbol{x}_t = [x_{1,t}, \cdots, x_{N,t}] \in [0, B]^N$, where $x_{it}$ represents the backlog of queue $i$ at time slot $t$ and $N$ is the number of queues. The decision is denoted by $\boldsymbol{y}_t = [y_{1,t}, \cdots, y_{N,t}]$ where $y_{it}$ represents the amount of traffic that is transmitted at time slot $t$. Similar to the assumptions in Section II, we assume that the immediate utility has the additive form of $u(\boldsymbol{x}, \boldsymbol{y}) = \sum_{i=1}^{N} u_i(x_i, y_i)$ where $u_i(x_i, y_i)$ represents the utility function of queue $i$, and the transmission cost is given by $c(h, \boldsymbol{y}) = c\left(h, \sum_{i=1}^{N} y_i\right)$. The immediate utility and transmission cost satisfy the following conditions:

Assumption 3: the utility function for each queue satisfies assumption 1;

Assumption 4: $c(h, y)$ is increasing and convex in $y$ for any give $h \in \mathcal{H}$.

From assumptions 3 and 4, we know that $u(\boldsymbol{x}, \boldsymbol{y}) - \lambda c(h, \boldsymbol{y})$ is supermodular in the pair of $(\boldsymbol{x}, \boldsymbol{y})$ and jointly concave in $(\boldsymbol{x}, \boldsymbol{y})$. Similar to the problem with one single queue, the following theorem shows that the optimal scheduling policy is also non-decreasing in the buffer length $\boldsymbol{x}$ for any given $h \in \mathcal{H}$ and the resulted post-decision state function is a concave function.

**Theorem** 6. With assumptions 3 and 4, the post-decision state function $V^{*,\lambda}(\boldsymbol{x}, h)$ is a concave function in $\boldsymbol{x}$ for any given $h \in \mathcal{H}$ and the optimal scheduling policy $\pi^{*,\lambda}(\boldsymbol{x}, h)$ is non-decreasing in $\boldsymbol{x}$ for any given $h \in \mathcal{H}$.

Proof: The proof is similar to the one in Theorem 2 and omitted here due to space limitations.

Similar to the approximation in the post-decision state-value function for the single queue problem, the concavity of the post-decision state function $V^{*,\lambda}(\boldsymbol{x}, h)$ in the backlog $\boldsymbol{x}$ enables us to approximate it using multi-dimensional piece-wise linear functions [26]. However, approximating a multi-dimensional concave function has high computation complexity and storage overhead due to the following reasons.

(i). To approximate an $N$-dimensional concave function, if we sample $m$ points in each dimension, the total number of samples to be evaluated is $m^N$. Hence, we need to update $m^N$ post-decision state-values in each time slot and store the $m^N$ post-decision state-values. We notice that the complexity still exponentially increases with the number of queues.

(ii). To evaluate the value at the post-decision states which are not the sample states, we require $N$-dimensional interpolation, which is often required to solve a linear programming problem [26]. Given the post-decision state-values at these sample points, computing the gap requires solving the linear program as well. Hence, the computation in solving the maximization for the state-value function update still remains complex.

However, we notice that, if the queues can be prioritized, this can significantly simplify the approximation complexity, as discussed next. First, we formally define the prioritized queues as follows.

**Definition (Priority queue)**: Queue $j$ has a higher priority than queue $k$ (denoted as $j \triangleleft k$) if the following condition holds:

$$u_j(x_j, y_j + \triangle y) - u_j(x_j, y_j) > u_k(x_k, y_k + \triangle y) - u_k(x_k, y_k), \forall x_j, x_k, y_j + \triangle y \leq x_j, y_k + \triangle y \leq x_k.$$

The priority definition in the above shows that transmitting the same amount of data from queue $j$ always gives us higher utility than transmitting data from queue $k$. One example is $u(\bm{x}, h, \bm{y}) = w_1 \min(x_1, y_1) + w_2 \min(x_2, y_2)$ with $w_1 = 1, w_2 = 0.8$. It is clear that queue 1 has higher priority than queue 2. In the following, we will show how the prioritization affects the packet scheduling policy and the state-value function representation. In the following, we assume that the $N$ queues are prioritized and $1 \triangleleft 2 \cdots \triangleleft N$. The following theorem shows that the optimal scheduling policy can be found queue by queue and the post-decision state-value function can be presented using $N$ one-dimensional concave functions.

**Theorem 7**: The optimal scheduling policy at the post-decision state $(\bm{x}, h)$ and post-decision state-value function can be solved as follows.

(i) The optimal scheduling policy for queue $i$ is obtained by solving the foresighted optimization:

$$y_i^* = \arg\max_{0 \leq y_i \leq \min(x_i + a_i, B)} \left\{ u_i(\min(x_i + a_i, B), y_i) - \lambda c\left(h, y_i + \sum_{j=1}^{i-1} y_j^*\right) + \alpha V_i^{*,\lambda}(\min(x_i + a_i, B) - y_i, h) \right\}, \forall i \quad (24)$$

(ii) The optimal scheduling policy satisfies the condition of $(x_i - y_i^*) y_j^* = 0$ if $i \triangleleft j$.

(iii) The post-decision state-value function $V_i^{*,\lambda}(x_i, h)$ is a one-dimensional concave function for fixed $h$ and is computed as

$$V_i^{*,\lambda}(x_i,h) = \sum_{a_1,\cdots,a_i} \sum_{h'} \prod_{j=1}^{i} p(a_j) p(h'|h) \cdot \max_{0 \leq y_i \leq \min(x_i+a_i,B)} \tag{25}$$

$$\left\{ \sum_{j=1}^{i-1} u_i\left(a_j, z_j^*\right) + u_i\left(\min(x_i + a_i, B), y_i\right) - \lambda c\left(h', y_i + \sum_{j=1}^{i-1} z_j^*\right) + \alpha V_i^{*,\lambda}\left(\min(x_i + a_i, B) - y_i, h'\right) \right\}, \forall i$$

where
$$z_i^* = \arg\max_{0 \leq z_i \leq a_i} \left\{ u_i(a_i, z_i) - \lambda c\left(h', z_i + \sum_{j=1}^{i-1} z_j^*\right) + \alpha V_i^{*,\lambda}(a_i - z_i, h') \right\}, \forall i.$$

Proof: See the Appendix F.

In Theorem 7, statements (i) and (ii) indicate that, when queue $i$ has a higher priority than queue $j$, the data in queue $i$ should be transmitted first before transmitting any data from queue $j$. In the other words, if $y_j^* > 0$ (i.e. some data are transmitted from queue $j$), then $x_i = y_i^*$ which means that all the data in queue $i$ has been transmitted. If $x_i > y_i^*$ (i.e. some data in queue $i$ are not transmitted yet), then $y_j^* = 0$ which means that there is no data transmitted from queue $j$. When transmitting the data from the lower priority queue, the optimal scheduling policy for this queue should be solved by considering the impacts of higher priority queues through the convex transmission cost, as shown in Eq. (24). We further notice that, in order to obtain the optimal scheduling policy, we only need to compute $N$ one-dimensional post-decision state-value functions each of which corresponds to one queue.

From the above discussion, we know that, since $i \triangleleft j$, the data in queue $i$ must be transmitted earlier than the data in queue $j$. Hence, to determine the optimal scheduling policy $y_i^*$, we only require the post-decision state-value function $V^{*,\lambda}((0,\cdots,0,x_i,\cdots,x_n),h)$. We further notice that, the data at the lower priority queues ($i \triangleleft j$) does not affect the scheduling policy for queue $i$. Statement (iii) indicates that, $V_i^{*,\lambda}(x_i,h)$ is updated by setting $x_k = 0, k \triangleleft i$. It is worth noting that the update of $V_i^{*,\lambda}(x_i,h)$ is one-dimensional optimization and $V_i^{*,\lambda}(x_i,h)$ is concave. Hence, we are able to develop online learning algorithms with adaptive approximation for updating $V_i^{*,\lambda}(x_i,h)$. The online learning algorithm is illustrated in Algorithm 2.

When compared to the priority queue systems [36] where there is no control on the amount of data to be transmitted at each time slot, our algorithm is similar in the transmission order, i.e. always transmitting the higher priority data first. However, our proposed method further determines how much should be transmitted at each priority queue at each time.

Algorithm 2: Online learning algorithm with adaptive approximation for transmission scheduling with multiple priority queues

---

**Initialize**: $\hat{V}_i^{1,\lambda}(\cdot, h) = 0, \forall i$ for all possible channel state $h \in \mathcal{H}$; post-decision state $s_0 = (\boldsymbol{x}_0, h_0)$ where $\boldsymbol{x}_0 = (x_{1,0}, \cdots, x_{N,0})$; $t = 1$.

**Repeat**:

    Observe the traffic arrival $\boldsymbol{a}_{t-1} = (a_{1,t-1}, \cdots, a_{N,t-1})$ and new channel state $h_t$;

    Compute the normal state $(\min(\boldsymbol{x}_{t-1} + \boldsymbol{a}_{t-1}, B), h_t)$;

    **For** $i = 1, \cdots, N$ // find the optimal scheduling policy

        Compute the optimal scheduling policy $y_{i,t}^*$ as in Eq. (24) by replacing $V_i$ with the estimated one $\hat{V}_i^{t-1,\lambda}$ and transmit the data.

    **End**

    **For** $i = 1, \cdots, N$ // update the post-decision state-value function

        Approximate the post-decision state-value function given by

$$\hat{V}_i^{t-1,\lambda}(x, h_{t-1}) = A_\delta\left((1-\beta_t)\hat{V}(x, h_{t-1}) + \beta_t J_i^{t-1,\lambda}(\min(x + a_{t-1}, B), h_t)\right);$$

        Where

$$J_i^{t,\lambda}(x, h) = \max_{0 \le y_i \le x} \left\{ \sum_{j=1}^{i-1} u_i(a_j, z_j^*) + u_i(\min(x + a_i, B), y_i) - \lambda c\left(h', y_i + \sum_{j=1}^{i-1} z_j^*\right) + \alpha \hat{V}_i^{t-1,\lambda}(\min(x + a_i, B) - y_i, h') \right\}.$$

        Compute $z_i^*$:

$$z_i^* = \arg\max_{0 \le z_i \le a_i} \left\{ u_i(a_i, z_i) - \lambda c\left(h_t, z_i + \sum_{j=1}^{i-1} z_j^*\right) + \alpha \hat{V}_i^{t-1,\lambda}(a_i - z_i, h_t) \right\}$$

    **End**

    Update the post-decision state $s_t = (\min(\boldsymbol{x}_{t-1} + \boldsymbol{a}_{t-1}, B) - \boldsymbol{y}_t^*, h_t)$;

    $t \leftarrow t + 1$;

**End**

---

## VI. SIMULATION RESULTS

In this section, we perform numerical simulations to highlight the performance of the proposed online learning algorithm with adaptive approximation and compare it with other representative scheduling solutions.

### A. Transmission scheduling with one queue

In this simulation, we consider a wireless user transmitting traffic data over a time-varying wireless channel. The objective is to minimize the average delay while satisfying the energy constraint. Due to Little's theorem [21], it is known that minimizing the average delay is equivalent to minimizing the average queue length (i.e. maximizing the negative queue length and $u(x,y) = -(x-y)$). The energy function for transmitting the amount of $y$ (in bits) traffic at the channel state $h$ is given by $c(h,y) = \frac{\sigma^2}{|h|^2}(2^y - 1)$, where $\sigma^2$ is the variance of the white Gaussian noise [21]. In this simulation, we choose $\bar{h}^2/\sigma^2 = 0.14$ where $\bar{h}$ is the average channel gain. We divide the entire channel gain range into eight regions each of which is represented by a representative state. The states are presented in Table 2. The incoming data is modelled as Poisson arrival [36] with an average arrival rate of 1.5Mbps. In order to obtain the average delay, we choose $\alpha = 0.95$. The transmission system is time-slotted with the time slot length of 10ms.

Table 2. Channel states used in the simulation

| Channel gain ($h^2/\sigma^2$) regions | Representative states |
|---|---|
| (0, 0.0280], (0.0280, 0.0580] (0.0580, 0.0960] (0.0960, 0.1400] (0.1400, 0.1980] (0.1980, 0.2780], (0.2780, 0.4160] (0.4160, $\infty$ ] | 0.0131, 0.0418, 0.0753, 0.1157, 0.1661, 0.2343, 0.3407, 0.6200 |

*A.1 Complexity of online learning with adaptive approximation*

In this simulation, we assume that the channel states transition is modelled as a finite-state Markov chain and the transition probability can be computed as in [20]. As we discussed in Section IV, by choosing different approximation error threshold $\delta$, we are able to approximate the post-decision state-value function by evaluating different number of states and at different accuracy. The simulation results are obtained by running the online learning algorithm for 10000 time slots. Figure 5 shows the delay-energy trade-off obtained by the online learning algorithms with different approximation error thresholds and Table 3 illustrates the corresponding number of states that need to be evaluated. It is easy to see that when the approximation error threshold $\delta$ increases from 0 to 30 (note that $\delta = 0$ indicates that no approximation is performed), the trade-off curve moves toward the upper-right corner which means that, in order to obtain the same delay, the learning algorithm with higher approximation error threshold increases the energy consumption. We notice that the energy increase is less than 5%. However, the number of states that are required to evaluate at each time slot is significantly reduced from 500 (corresponding the buffer size $B = 500$) to 5.

In order to further reduce the computation complexity, instead of updating the post-decision state-value function every time slot as performed in the above, we update the post-decision state-value function every $T$ time slots where $T = 1,5,10,20,30,40$. The delay-energy trade-offs obtained by the online learning algorithm with adaptive approximation are depicted in Figure 6 where $\delta =10$. On one hand, we note that, when $T$ increases from 1 to 40, the amount of energy consumed in order to achieve the same delay performance is increased. However, the increase is less than 10%. On the other hand, from Table 3, we note that we only need to update 10 states at each time slot when $\delta =10$. If we update the post-decision state-value function every $T = 40$, then on average, we only need to update 1.25 states per time slot which significantly reduces the learning complexity.

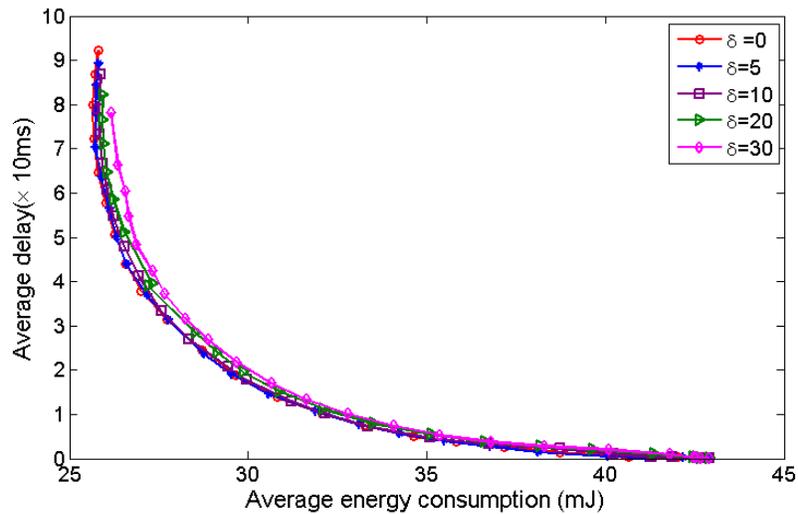

Figure 5.    Delay-energy trade-off obtained by online learning algorithm with different approximation error thresholds

Table 3. Number of states that are updated at each time slot

|  | $\delta = 0$ | $\delta = 5$ | $\delta = 10$ | $\delta = 20$ | $\delta = 30$ |
|---|---|---|---|---|---|
| #states updated per time slot | 500 | 14 | 10 | 7 | 5 |

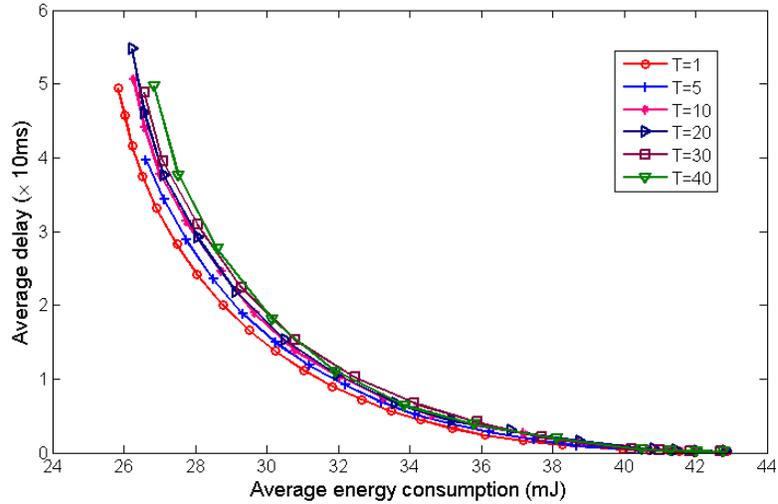

Figure 6.  Delay-energy trade-off obtained by online learning algorithm with different update frequencies

## A.2  *Comparison with other representative methods*

In this section, we compare our proposed online learning algorithm with other representative methods. Specially, we first compare our method with the stability-constrained optimization method proposed in [13] for single-user transmission. We consider three scenarios: (i) i.i.d. channel gain which is often assumed by the stability-constrained optimization; (ii) Markovian channel gain which is assumed in this paper; (iii). Non-Markovian stationary channel gain (generated by moving averaging model [34]). In this simulation, the trade-off parameter (Lagrangian multiplier) $\lambda$ is updated via virtual queue in the stability-constrained optimization and via stochastic subgradient method as shown in Eq. (13) in our proposed method. In our method, $\delta = 10$ and $T = 10$.

Figure 7 to Figure 9 show the delay-energy consumption trade-offs when the data is transmitted over these three different channels. From these figures, we note that our proposed method outperforms the stability-constrained optimization at both the large delay region ($\geq 15$) and the small delay region. We also note that, in the large delay region, the difference between our method and the stability-constrained optimization becomes small since the stability-constrained optimization method asymptotically achieves the optimal energy consumption and our method is $\varepsilon$-optimal. However, in the small delay region, our method can significantly reduce the energy consumption for the same delay performance. We further notice that the stability-constrained method could not achieve zero delay (i.e. the incoming data is process once it enters into the queue) even if the energy consumption increases. This is because the stability-constrained optimization method only minimizes the energy consumption at the large delay region and does not perform optimal energy allocation at the small delay region since the queue length is small. In contrast, our proposed online learning is able to take care of both regions by adaptively approximating the post-decision state-value functions.

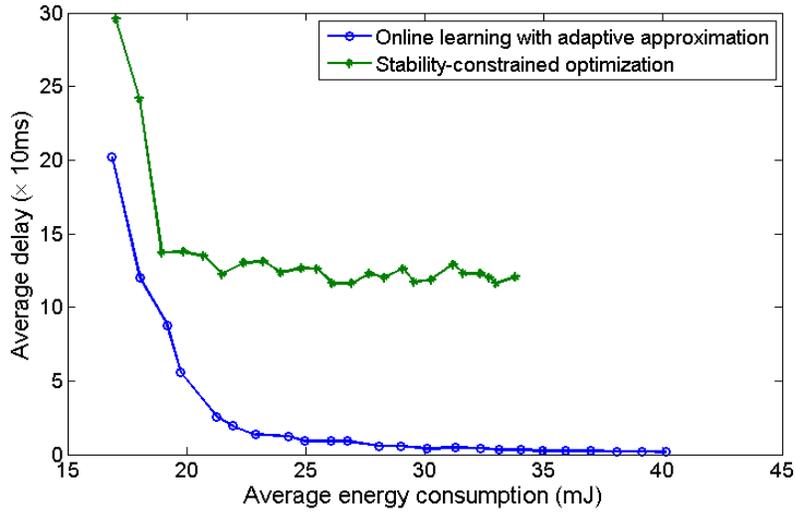

Figure 7.    Delay-energy trade-off when the underlying channel is i.i.d.

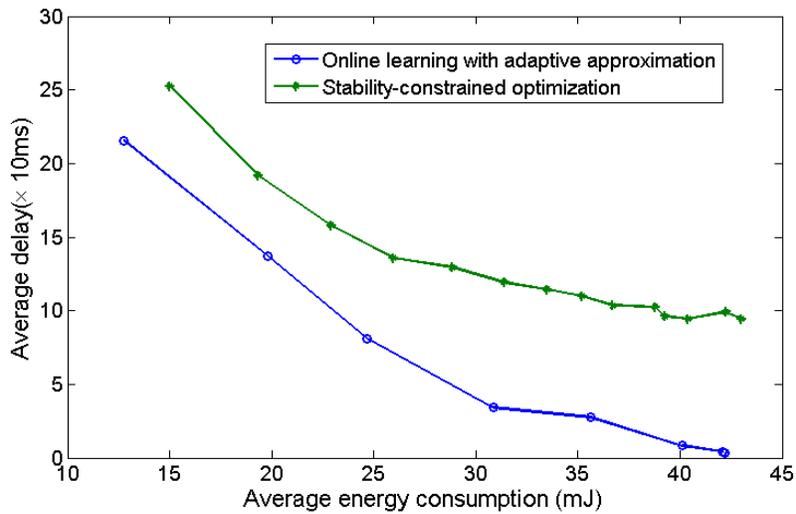

Figure 8.    Delay-energy trade-off when the underlying channel is Markovian

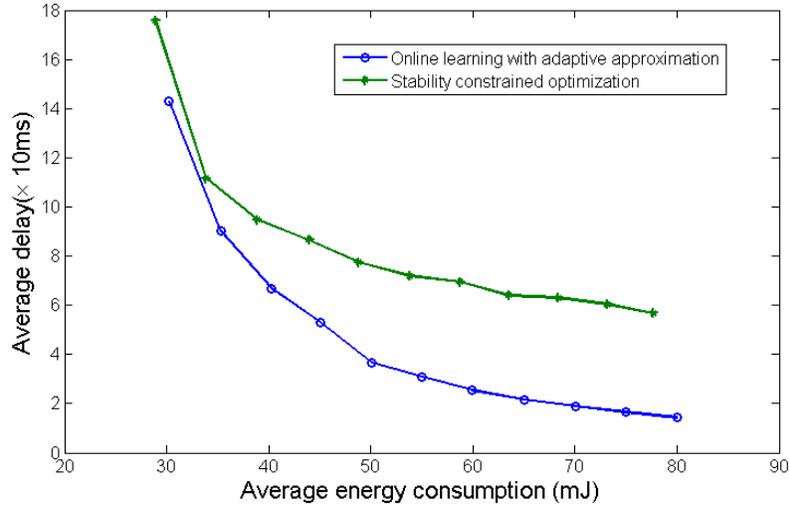

Figure 9. Delay-energy trade-off when the underlying channel is non-Markovian

We then compare our proposed method with Q-learning algorithm proposed in [17]. In this simulation, we transmit the data over the Markovian channel. In the Q-learning algorithm, the post-decision state-value function is updated for one state per time slot. Figure 10 shows the delay-energy trade-offs. The delay-energy trade-off of our proposed method is obtained by running our method for 5000 time slots. The delay-energy trade-off of the Q-learning algorithm is obtained by running Q-learning algorithm for 50000 time slots. It can be seen from Figure 10 that our proposed method outperforms the Q-learning even when our algorithm learns only over 5000 time slots and the Q-learning algorithm learns over 50000 time slot. Hence, our method significantly reduces the amount of time to learn the underlying dynamics (i.e. experiencing faster learning rate) comparing to the Q-learning algorithm.

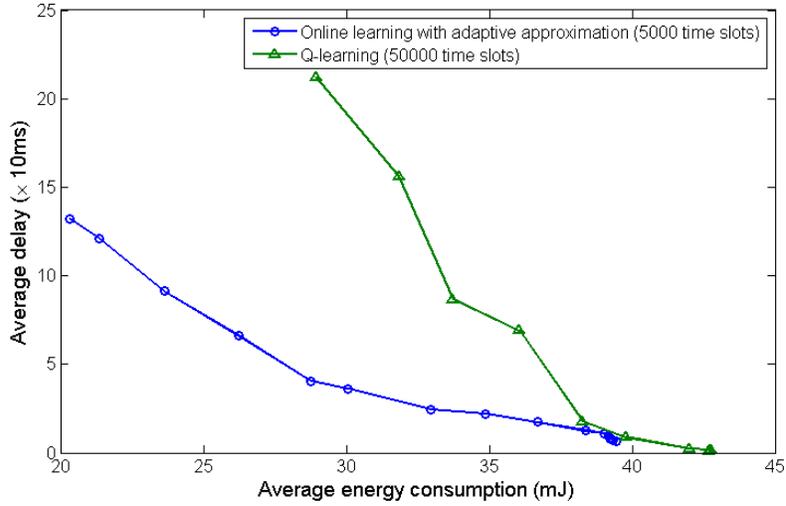

Figure 10.  Delay-energy trade-off obtained by different online learning algorithms when the channel is Markovian

## B. Transmission scheduling with multiple priority queues

In this section, we consider that the wireless user schedules the prioritized data over a time-varying wireless channel. The channel configuration is the same as in Section VI.A. The wireless user has two prioritized classes of data to be transmitted. The utility function is given by $u(\boldsymbol{x},h,\boldsymbol{y}) = w_1 \min(x_1, y_1) + w_2 \min(x_2, y_2)$ where $w_1 = 1.0$ and $w_2 = 0.8$ represent the importance of the data at classes 1 and 2, respectively. Thus, we have $1 \triangleleft 2$. Figure 11 illustrates the utility-energy trade-offs obtained by the proposed online learning algorithm and the stability-constrained optimization method. Figure 12 shows the corresponding delay-energy trade-offs experienced by each class of data. It can be seen from Figure 11 that, at the same energy consumption, our proposed algorithm can achieve an utility 2.2 times higher of the one obtained by the stability-constrained optimization method. It is worth noting that class 1 has less delay than class 2 which is demonstrated in Figure 12, because class 1 has higher priority. It can also be seen from Figure 12 that, the delay is reduced by 50%, on average, for each class in our method when compared to the stability-constrained optimization method. This improvement is due to the fact that our proposed method explicitly considers the time-correlation in the channel state transition and the priorities in the data.

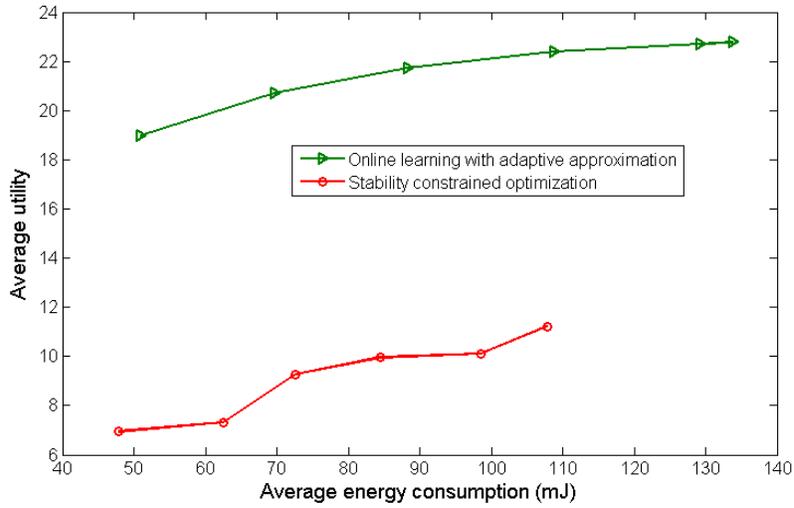

Figure 11.  Utility-energy trade-off of prioritized traffic transmission by different online methods when the channel is

Markovian

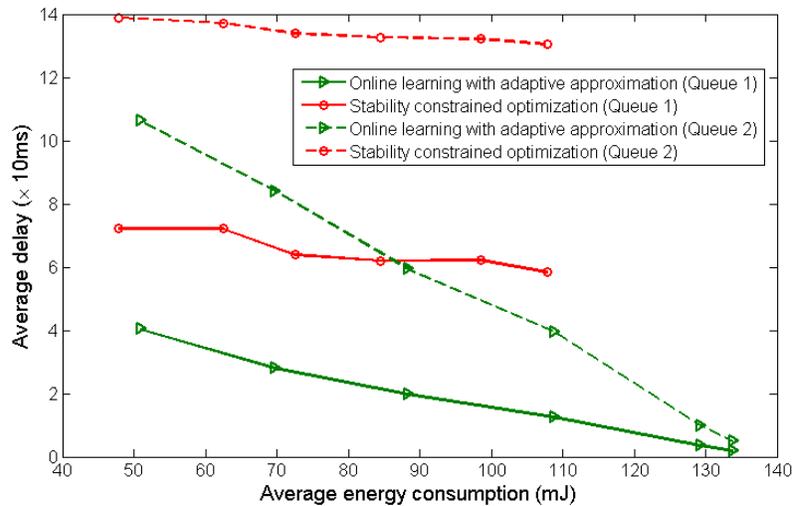

Figure 12.  Delay-energy trade-off of each class in prioritized traffic transmission by different online methods when the

channel is Markovian

## VII.  CONCLUSIONS

In this paper, we first establish the structural results of the optimal solutions to the constrained MDP formulation of the transmission scheduling problems. Based on these structural properties, we propose to adaptively approximate the post-decision state-value function using piece-wise linear functions which can preserve the structural properties. Furthermore, this approximation allows us to compactly represent the post-decision state-value functions and learn them with low complexity. We prove that the online learning with adaptive approximation converges to the $\varepsilon$-optimal solutions the size of which is controlled by the pre-determined approximation error. We extend our method to the heterogeneous data transmission in which the incoming traffic is prioritized. An extension of our method considers heterogeneous data transmission in which the data has different delay-deadlines, priorities and dependencies has been discussed in [29]. Another possible extension is that multi-user data transmission in which the users share the same network resources. The interesting issue here is how the network resource can be dynamically allocated and how the users learn their own state-value functions. Partial results on this have been presented in [33].

We notice that the method presented in this paper can be applied to other classes of applications in which the immediate utilities are supermodular, and where the decisions need to be adapted dynamically, over time, and which operate in unknown environments. Examples of such applications are cross-layer optimization [33], adaptive media encoding/decoding [31], dynamic resource allocation for large-scale data center [32] and streaming mining systems [30], etc.

## Appendix

*Appendix A: Proof of Theorem 2*

We use backward induction to prove this theorem. Since the value iteration converges to the optimal post-decision state-value function $V^{*,\lambda}(x,h)$ for any initial post-decision state-value function $V_0^\lambda(x,h)$. We choose $V_0^\lambda(x,h)$ to be monotonic and concave in $x$ for any $h \in \mathcal{H}$. Due to the symmetry, we only consider the case that $u(x,y)$ is decreasing in $x$ but supermodular in $(x,y)$. Then, we choose $V_0^\lambda(x,h)$ to be non-increasing and concave in $x$, e.g. $V_0^\lambda(x,h) = -x$.

Now assume that $V_{m-1}^\lambda(x,h)$ is concave in $x$ for any $h \in \mathcal{H}$, and $V_m^\lambda(x,h), m = 1,2,3,\cdots$ are computed as

$$V_m^\lambda(x,h) = \sum_a \sum_{h' \in \mathcal{H}} p(a) p(h' \mid h) \max_{0 \leq y \leq \min(x+a,B)} \left[ u(\min(x+a,B),y) - \lambda c(h',y) + \alpha V_{m-1}^\lambda(\min(x+a,B) - y, h') \right].$$

It can be proved that $\lim_{m \to \infty} V_m^\lambda(x,h) = V^{*,\lambda}(x,h)$. In the following, we only need to prove that $V_m^\lambda(x,h)$ is concave. We also notice that and the optimal scheduling policy $\pi(x,h)$ can be obtained as $\pi^{*,\lambda}(x,h) = \lim_{m \to \infty} \pi_m^\lambda(x,h)$ where

$$\pi_m^\lambda(x,h) = \arg \max_{0 \leq y \leq x} Q_m^\lambda(x,h,y)$$

and

$$Q_m^\lambda(x,h,y) = \left[ u(x,y) - \lambda c(h,y) + \alpha V_{m-1}^\lambda(x-y,h) \right].$$

To prove that $\pi^{*,\lambda}(x,h)$ is increasing in $x$ for any $h$, we only need to prove that $Q_m^\lambda(x,h,y)$ is supermodular in $(x,y)$. First, we note that $V_{m-1}^\lambda(x,h)$ is concave in $x$ by our assumption. Then it holds that

$$V_{m-1}^\lambda(x_1,h) + V_{m-1}^\lambda(x_2,h) \leq V_{m-1}^\lambda((1-\eta)x_1 + \eta x_2, h) + V_{m-1}^\lambda(\eta x_1 + (1-\eta)x_2, h), \forall \eta \in [0,1].$$

Consider that $x' \geq x$ and $y' \geq y$ and $x \geq y'$. Let $x_1 = x' - y$ and $x_2 = x - y'$ and $\eta = \dfrac{y' - y}{x' - x + y' - y}$. Then

$$(1-\eta)x_1 + \eta x_2 = (x' - y'), \eta x_1 + (1-\eta)x_2 = (x - y).$$

Hence, we have

$$V_{m-1}^\lambda(x' - y, h) + V_{m-1}^\lambda(x - y', h) \leq V_{m-1}^\lambda(x - y, h) + V_{m-1}^\lambda(x' - y', h).$$

By rearranging, we obtain

$$V_{m-1}^\lambda(x - y', h) - V_{m-1}^\lambda(x - y, h) \leq V_{m-1}^\lambda(x' - y', h) - V_{m-1}^\lambda(x' - y, h)$$

which proves that $V_{m-1}^\lambda(x-y,h)$ is supermodular in $(x,y)$. It turns out that $\pi_m^\lambda(x,h)$ is increasing in $x$ for any $h$ since $Q_m^\lambda(x,h,y)$ is supermodular. Let $m \to \infty$, we know $\pi^{*,\lambda}(x,h)$ is also increasing in $x$ for any $h$.

Next, we try to prove that $V_m^\lambda(x,h)$ is concave in $x$ for any $h$. We first prove that

$$J_m^\lambda(x,h) = \max_{0 \leq y \leq x} Q_m^\lambda(x,h,y)$$

is concave in $x$ for any $h$. For any $x_1, x_2$, we assume that the optimal scheduling are $y_1 = y^*(x_1, h)$ and $y_2 = y^*(x_2, h)$, respectively. $\forall \eta \in [0,1]$, we have

$$J_m^\lambda(x_1, h) + J_m^\lambda(x_2, h) = u(x_1, y_1) - \lambda c(h, y_1) + \alpha V_{m-1}^\lambda(x_1 - y_1, h)$$
$$+ u(x_2, y_2) - \lambda c(h, y_2) + \alpha V_{m-1}^\lambda(x_2 - y_2, h)$$
$$\leq u(\eta x_1 + (1-\eta) x_2, \eta y_1 + (1-\eta) y_2) - \lambda c(h, \eta y_1 + (1-\eta) y_2) +$$
$$\alpha V_{m-1}^\lambda(\eta(x_1 - y_1) + (1-\eta)(x_2 - y_2), h)$$
$$= Q_m^\lambda(\eta x_1 + (1-\eta) x_2, h, \eta y_1 + (1-\eta) y_2)$$
$$\leq J_m^\lambda(\eta x_1 + (1-\eta) x_2, h)$$

The first inequality is from the facts that $u(x,y)$ is jointly concave in $(x,y)$ and $c(h, y_1)$ is convex in $y$ and $V_{m-1}^\lambda(x,h)$ is concave in $x$. The second inequality is from the fact that $J_m^\lambda(x,h) = \max_{0 \leq y \leq x} Q_m^\lambda(x,h,y)$.

Then $J_m^\lambda(x,h)$ is concave in $x$. We further define $\phi(x,a) = \min(B-x, a)$. We note that $V_m^\lambda(x,h) = \sum_{\phi(x,a)} \sum_{h' \in \mathcal{H}} p(\phi(x,a)) p(h' \mid h) J_m^\lambda(x + \phi(x,a), h')$ and $p(\phi(x,a))$ is uniquely determined by the distribution of $p(a)$ and $x$. Then $\tilde{J}_m^\lambda(\tilde{s})$ is concave as well. ∎

*Appendix B: Approximating the concave function*

In this section, we present a method to approximate a one-dimensional concave function. Considering a concave and increasing function $f : [a,b] \to \mathbb{R}$ with $n$ points $\{(x_i, f(x_i)) \mid i = 1, \cdots, n\}$ and $x_1 < x_2 < \cdots < x_n$. Based on these $n$ points, we are able to give the lower and upper bounds on the function $f$. It is well-known that the straight line through the points $(x_i, f(x_i))$ and $(x_{i+1}, f(x_{i+1}))$ for $i = 1, \cdots, n-1$ is the lower bound of the function $f(x)$ for $x \in [x_i, x_{i+1}]$. It is also well-known that the straight lines through the points $(x_{i-1}, f(x_{i-1}))$ and $(x_i, f(x_i))$ for $i = 2, \cdots, n$ and the points $(x_{i+1}, f(x_{i+1}))$ and $(x_{i+2}, f(x_{i+2}))$ for $i = 1, \cdots, n-2$ are the upper bounds of the function $f(x)$ for $x \in [x_i, x_{i+1}]$. This is illustrated in Figure 13.

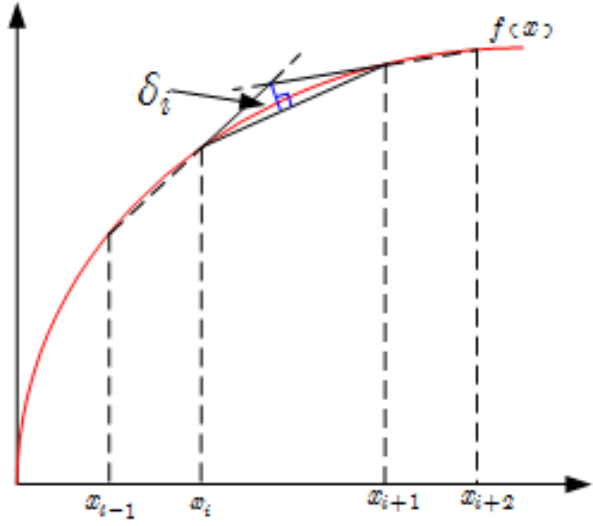

Figure 13. Lower and upper bound of the concave function $f(x)$ in the range of $[x_i, x_{i+1}]$.

This idea can be summarized in the following lemma.

**Lemma**: Given $n$ points $\{(x_i, f(x_i)) \mid i = 1, \cdots, n\}$ with $x_1 = a < x_2 < \cdots < x_n = b$ and $f(x)$ is an concave and increasing function, then

(i) the piece-wise linear function $\hat{f}(x) = k_i x + b_i$ if $x_i \leq x \leq x_{i+1}$ is the lower bound of $f(x)$ where

$$k_i = \frac{f(x_{i+1}) - f(x_i)}{x_{i+1} - x_i}, b_i = \frac{x_{i+1} f(x_i) - x_i f(x_{i+1})}{x_{i+1} - x_i}.$$

(ii) The maximum gap between the piece-wise linear function $\hat{f}(x)$ and $f(x)$ is given by

$$\delta = \max_{i=1,\cdots,n-1} \delta_i \qquad (26)$$

where

$$\delta_i = \begin{cases} \dfrac{\|k_1 x_1 + b_1 - k_2 x_1 - b_2\|}{\sqrt{1 + k_1^2}} & i = 1 \\[2ex] \dfrac{\left\|\dfrac{k_{i-1} - k_i}{k_{i-1} - k_{i+1}}(b_{i-1} - b_{i+1}) - (b_{i-1} - b_i)\right\|}{\sqrt{1 + k_i^2}} & 1 < i < n-1 \\[2ex] \dfrac{\|k_{n-1} x_n + b_{n-1} - k_{n-2} x_n - b_{n-2}\|}{\sqrt{1 + k_{n-1}^2}} & i = n-1 \end{cases} \qquad (27)$$

Proof: The proof can be easily shown based on Figure 1 and basic algebra geometry knowledge. We omit the proof here for space limitations.

In the following, we present an iterative method to build the lower bound piece-wise linear function $\hat{f}(x)$ with the pre-determined approximation threshold $\delta$. This iterative method is referred to as the sandwich algorithm in the literature [25].

The lower bound piece-wise linear function and corresponding gap are generated in an iterative way. We start evaluating the concave function $f(x)$ at the boundary point $x = a$ and $x = b$, i.e. $I = \{[a,b]\}, n = 2$. Then we can obtain the piece-wise linear function $\hat{f}_0(x)$ with the maximum gap of $\delta^0 = \|f(b) - f(a)\|$. Assuming that, at iteration $k$, the maximum gap is $\delta^k$ which is computed at the corresponding interval $\left[x_{j^k}, x_{j^k+1}\right]$. If the gap $\delta^k > \delta$, we evaluate the function $f(x)$ at the additional point $y = \left(x_{j^k} + x_{j^k+1}\right)/2$. We partition the interval $\left[x_{j^k}, x_{j^k+1}\right]$ into the two intervals $\left[x_{j^k}, y\right]$ and $\left[y, x_{j^k+1}\right]$. We further evaluate the gaps for the intervals $\left[x_{j^k-1}, x_{j^k}\right], \left[x_{j^k}, y\right], \left[y, x_{j^k+1}\right]$, and $\left[x_{j^k+1}, x_{j^k+2}\right]$ using Eq. (27). The maximum gap is then updated. We repeat this procedure until the maximum gap is less than the given approximation threshold $\delta$. The procedure is summarized in Algorithm 3.

Algorithm 3. Sandwich algorithm for approximating the concave function

**Initialize**: $x_1^0 = a$, $x_2^0 = b$, $f(x_1^0)$, $f(x_2^0)$, $\delta^0 = f(x_2^0) - f(x_1^0)$, $j^0 = 1$, $k = 0$ and $n = 2$;
**Repeat**:
    $y = \left(x_{j^k} + x_{j^k+1}\right)/2$; Compute $f(y)$;
    Partition the interval $\left[x_{j^k}, x_{j^k+1}\right]$ into $\left[x_{j^k}, y\right]$ and $\left[y, x_{j^k+1}\right]$.
    Compute the gaps corresponding to the intervals $\left[x_{j^k-1}, x_{j^k}\right], \left[x_{j^k}, y\right], \left[y, x_{j^k+1}\right]$ and $\left[x_{j^k+1}, x_{j^k+2}\right]$.
    $x_{j+1}^{k+1} \leftarrow x_j^k$ for $j = j^k + 1, \cdots, n$; $x_{j^k}^{k+1} \leftarrow y$; $x_j^{k+1} \leftarrow x_j^k$ for $j = 1, \cdots, j^k$;
    $k \leftarrow k + 1$; $n \leftarrow n + 1$;
    Update the maximum gap $\delta^k$ and the index $j^k$ corresponding to the interval having the maximum gap.
**Until** $\delta^k \leq \delta$.

This algorithm allows us to adaptively select the points $\{x_1, \cdots, x_{n_\delta}\}$ to evaluate the value of $f(x)$ based on the pre-determined threshold $\delta$. This iterative method provides us a simple way to approximate the post-decision state-value function which is concave in the backlog $x$.

*Appendix C: Proof of Theorem 3*

Proof: We first note that $V^{*,\lambda}(x,h) = \lim_{m \to \infty} V_m^\lambda(x,h) = T^\infty V_0^\lambda(x,h)$. Since $T$ is a maximum norm $\alpha$-contractor, we have $\|TV - TV'\|_\infty \leq \alpha \|V - V'\|_\infty$. If $0 \leq V(x,h) - V'(x,h) \leq \delta$, then $TV - TV' \leq \alpha\delta$ The piece-wise linear approximation operator has $0 \leq TV - A_\delta TV \leq \delta$. Then we have

$$\|T^m V_0^\lambda - (A_\delta T)^m V_0^\lambda\|_\infty = \|TT^{m-1}V_0^\lambda - A_\delta T(A_\delta T)^{m-1}V_0^\lambda\|_\infty \leq \delta + \alpha\|T^{m-1}V_0^\lambda - (A_\delta T)^{m-1}V_0^\lambda\|_\infty$$
$$\leq \delta + \alpha\left(\delta + \alpha\|T^{m-2}V_0^\lambda - (A_\delta T)^{m-2}V_0^\lambda\|_\infty\right) \leq ... \leq \delta \sum_{i=1}^{m} \alpha^{i-1}.$$

Let $m \to \infty$, we have $\|T^\infty V_0^\lambda - (A_\delta T)^\infty V_0^\lambda\|_\infty \leq \delta \sum_{i=1}^{\infty} \alpha^{i-1} = \frac{\delta}{1-\alpha}$. It can be easily shown that $T^\infty V_0^\lambda \geq (A_\delta T)^\infty V_0^\lambda$. Hence, we have $0 \leq T^\infty V_0^\lambda - (A_\delta T)^\infty V_0^\lambda \leq \frac{\delta}{1-\alpha}$. This proves statement (i).

For any piece-wise linear function $V_0^\lambda$ and $V_0'^\lambda$, we have $0 \leq T^\infty V_0^\lambda - (A_\delta T)^\infty V_0^\lambda \leq \frac{\delta}{1-\alpha}$ and $0 \leq T^\infty V_0'^\lambda - (A_\delta T)^\infty V_0'^\lambda \leq \frac{\delta}{1-\alpha}$. We can conclude that $0 \leq \|(A_\delta T)^\infty V_0'^\lambda - (A_\delta T)^\infty V_0^\lambda\|_\infty \leq \frac{\delta}{1-\alpha}$, which proves statement (ii). ∎

*Appendix D: Proof of Theorem 4*

Proof: To prove this, we define the foresighted optimization operator as follows.

$$T_{a,h}V(x,h) = \max_{0 \leq y \leq \min(x+a,B)} [u(\min(x+a,B),y) - \lambda c(h,y) + \alpha V(\min(x+a,B) - y, h)].$$

Then the post-decision state-based Bellman equations can be rewritten as

$$V^{*,\lambda} = \mathop{\mathbf{E}}_{a,h} T_{a,h} V^{*,\lambda}$$

where $\mathbf{E}$ is the expectation over the data arrival and channel state transition and the operator is a maximum norm $\alpha$-contraction.

The online learning of the post-decision state-value function in Eq. (21) can be re-expressed by

$$V^{t,\lambda} = V^{t-1,\lambda} + \beta_t \left(T_{a,h} V^{t-1,\lambda} - V^{t-1,\lambda}\right).$$

Similar to [24], it can be shown that the convergence of the online learning algorithm is equivalent to the convergence of the following O.D.E.:

$$\dot{V}^\lambda = \mathop{\mathbf{E}}_{a,h} T_{a,h} V^\lambda - V^\lambda.$$

Since $T_{a,h}$ is a contraction mapping, the asymptotic stability of the unique equilibrium point of the above O.D.E. is guaranteed [24]. This unique equilibrium point corresponds to the optimal post-decision state-value function $V^{*,\lambda}$. ∎

*Appendix E: Proof of Theorem 5*

Proof: From the proof of Theorem 4, we know that, the online learning algorithm with the adaptive approximation can be re-expressed as

$$V^{t,\lambda} = A_\delta \left( V^{t-1,\lambda} + \beta_t \left( T_{a,h} V^{t-1,\lambda} - V^{t-1,\lambda} \right) \right).$$

The corresponding O.D.E. is

$$\dot{V}^\lambda = A_\delta \left( \mathop{\mathbf{E}}_{a,h} T_{a,h} V^\lambda \right) - V^\lambda.$$

By the contraction mapping and the property of $A_\delta$, we can show that $\left\| V^{*,\lambda} - \hat{V}^{*,\lambda} \right\|_\infty \leq \dfrac{\delta}{1-\alpha}$. ∎

*Appendix F: Proof of Theorem 7*

Proof: We prove this by backward induction. We choose the initial post-decision state value function $V^{0,\lambda}(\boldsymbol{x},h) = 0$. Then,

$$\boldsymbol{y}^{1,\lambda} = \arg \max_{0 \leq \boldsymbol{y} \leq \boldsymbol{x}} \left\{ \sum_{i=1}^N u_i(x_i, y_i) - \lambda c \left( h, \sum_{i=1}^N y_i \right) \right\}.$$

Due to the priorities $1 \lhd 2 \cdots \lhd N$, we know that

$$y_i^{1,\lambda} = \arg \max_{0 \leq y_i \leq x_i} \left\{ u_i(x_i, y_i) - \lambda c \left( h, \sum_{j=1}^{i-1} y_j^{1,\lambda} + y_i \right) \right\}.$$

It can be shown that $\left( x_i - y_i^{1,\lambda} \right) y_j^{1,\lambda} = 0$, if $i \lhd j$.

We define

$$J_i^{1,\lambda}((x_1,\cdots,x_i),h) = \max_{0 \leq y_j \leq x_j} \left\{ \sum_{j=1}^i u_j(x_j, y_j) - \lambda c \left( h, \sum_{j=1}^i y_j \right) \right\}.$$

$J_i^{1,\lambda}$ is the state-value function for the normal state corresponding to the queues $1,\cdots,i$. It can be shown that $J_i^{1,\lambda}$ is a concave function.

We further notice that $J_i^{1,\lambda}((x_1,\cdots,x_i-\Delta y),h)-J_j^{1,\lambda}((x_1,\cdots,x_j-\Delta y),h)\geq u_j(x_j,\Delta y)-u_i(x_i,\Delta y)$ if $i\lhd j$.

The post-decision state-value function for queue $i$ is computed by

$$V_i^{1,\lambda}(x_i,h)=\sum_{a_1,\cdots,a_i}\prod_{j=1}^i p_j(a_j)\sum_{h'}p(h'\mid h)J_i^{1,\lambda}((a_1,\cdots,a_{i-1},x_i+a_i),h').$$

Hence, $V_i^{1,\lambda}(x_i-\Delta y,h)-V_j^{1,\lambda}(x_j-\Delta y,h)\geq u_j(x_j,\Delta y)-u_i(x_i,\Delta y)$ as well if $i\lhd j$.

Let us assume that $V_i^{m-1,\lambda}(x_i-\Delta y,h)-V_j^{m-1,\lambda}(x_j-\Delta y,h)\geq u_j(x_j,\Delta y)-u_i(x_i,\Delta y)$. Since the lower priority data will not affect the transmission of the higher priority data, we have

$$\left[u_i(x_i,y_i+\Delta y)+\sum_{k=1,k\neq i}^n u_k(x_k,y_k)-\lambda c\left(h,\sum_{k=1}^n y_k+\Delta y\right)+\alpha V^{m-1,\lambda}((x_i-y_i-\Delta y,\boldsymbol{x}_{-i}-\boldsymbol{y}_{-i}),h)\right]$$

$$-\left[u_j(x_j,y_j+\Delta y)+\sum_{k=1,k\neq j}^n u_k(x_k,y_k)-\lambda c\left(h,\sum_{k=1}^n y_k+\Delta y\right)+\alpha V^{m-1,\lambda}((x_j-y_j-\Delta y,\boldsymbol{x}_{-j}-\boldsymbol{y}_{-j}),h)\right]$$

$$=u_i(x_i,y_i+\Delta y)-u_j(x_j,y_j+\Delta y)+\alpha\left(V^{m-1,\lambda}((x_i-y_i-\Delta y,\boldsymbol{x}_{-i}-\boldsymbol{y}_{-i}),h)-V^{m-1,\lambda}((x_j-y_j-\Delta y,\boldsymbol{x}_{-j}-\boldsymbol{y}_{-j}),h)\right)$$

$$=u_i(x_i,y_i+\Delta y)-u_j(x_j,y_j+\Delta y)+\alpha\left(V_i^{m-1,\lambda}(x_i-y_i-\Delta y,h)-V_j^{m-1,\lambda}(x_j-y_j-\Delta y,h)\right)$$

$$\geq u_i(x_i,y_i+\Delta y)-u_j(x_j,y_j+\Delta y)+\alpha\left(u_j(x_j,y_j+\Delta y)-u_i(x_i,y_i+\Delta y)\right)>0$$

This means that, if $i\lhd j$, then transmitting the data from queue $i$ results in higher utility than transmitting the data from queue $j$. Hence, the optimal packet scheduling satisfies $(x_i-y_i^{m,\lambda})y_j^{m,\lambda}=0$.

We further notice that

$$y_i^{m,\lambda}=\arg\max_{0\leq y_i\leq x_i}\left\{u_i(x_i,y_i)-\lambda c\left(h,\sum_{j=1}^{i-1}y_j^{m,\lambda}+y_i\right)+\alpha V_i^{m-1,\lambda}(x_i-y_i,h)\right\},$$

$$J_i^{m,\lambda}((x_1,\cdots,x_i),h)=\max_{0\leq y_j\leq x_j}\left\{\sum_{j=1}^i u_j(x_j,y_j)-\lambda c\left(h,\sum_{j=1}^i y_j\right)+\alpha V_i^{m-1,\lambda}(x_i-y_i,h)\right\}$$

and

$$V_i^{m,\lambda}(x_i,h)=\sum_{a_1,\cdots,a_i}\prod_{j=1}^i p_j(a_j)\sum_{h'}p(h'\mid h)J_i^{m,\lambda}((a_1,\cdots,a_{i-1},x_i+a_i),h').$$

It can also be shown that $J_i^{m,\lambda}((x_1,\cdots,x_i-\triangle y),h) - J_j^{m,\lambda}((x_1,\cdots,x_j-\triangle y),h) \geq u_j(x_j,\triangle y) - u_i(x_i,\triangle y)$ and

$V_i^{m-1,\lambda}(x_i-\triangle y,h) - V_j^{m-1,\lambda}(x_j-\triangle y,h) \geq u_j(x_j,\triangle y) - u_i(x_i,\triangle y)$.

Let $m \to \infty$, we have $y_i^{m,\lambda} \to y_i^{*,\lambda}$ and $V_i^{m,\lambda}(x_i,h) \to V_i^{*,\lambda}(x_i,h)$, which proves Theorem 7. ∎

## REFERENCES

[1] R. Berry and R. G. Gallager, "Communications over fading channels with delay constraints," *IEEE Trans. Inf. Theory*, vol 48, no. 5, pp. 1135-1149, May 2002.

[2] A. Fu, E. Modiano, and J. Tsitsiklis. "Optimal energy allocation for delay-constrained data transmission over a time-varying channel," *IEEE Proceedings of INFOCOM*, 2003.

[3] W. Chen, M. J. Neely, and U. Mitra, "Energy-efficient transmission with individual packet delay constraints," *IEEE Trans. Inform. Theory*, vol. 54, no. 5, pp. 2090-2109, May. 2008.

[4] E. Uysal-Biyikoglu, B. Prabhakar, and A. El Gamal, "Energy-efficient packet transmission over a wireless link," *IEEE/ACM Trans. Netw.*, vol. 10, no. 4, pp. 487-499, Aug. 2002.

[5] M. Goyal, A. Kumar, and V. Sharma, "Optimal cross-layer scheduling of transmissions over a fading multiacess channel," *IEEE Trans. Info. Theory*, vol. 54, no. 8, pp. 3518-3536, Aug. 2008.

[6] M. Agarwal, V. Borkar, and A. Karandikar, "Structural properties of optimal transmission policies over a randomly varying channel," *IEEE Transactions on Automatical Control*, vol 53, no. 6, pp. 1476-1491, July 2008.

[7] D. Djonin and V. Krishnamurthy, "Structural results on optimal transmission scheduling over dynamical fading channels: A constrained Markov decision process Approach," in Wireless Communications, Editor: G. Yin, Institute for Mathematics and Applications (IMA) ,Springer Verlag 2006.

[8] T. Holliday, A. Goldsmith, and P. Glynn, "Optimal power control and source-channel coding for delay constrained traffic over wireless channels," *Proceedings of IEEE International Conference on Communications*, vol. 2, pp. 831 - 835, May 2002.

[9] A. Jalali, R. Padovani, and R. Pankaj, "Data throughput of cdma-hdr a high efficiency data rate personal communication wireless system," *IEEE Vehicular Technology Conference*, May 2000.

[10] L. Tassiulas and A. Ephremides, "Stability properties of constrained queueing systems and scheduling policies for maximum throughput in multihop radio networks," *IEEE Transacations on Automatic Control*, vol. 37, no. 12, pp. 1936-1949, Dec. 1992.

[11] P.R. Kumar and S.P. Meyn, "Stability of queueing networks and scheduling policies*,*" *IEEE Trans. on Automatic Control*, Feb. 1995.

[12] A. Stolyar, "Maximizing queueing network utility subject to stability: greedy primal-dual algorithm," *Queueing Systems*, vol. 50, pp. 401-457, 2005.

[13] L. Georgiadis, M. J. Neely, and L. Tassiulas, "Resource allocation and cross-layer control in wireless networks," *Foundations and Trends in Networking*, vol. 1, no. 1, pp. 1-149, 2006.

[14] P. Chou, and Z. Miao, "Rate-distortion optimized streaming of packetized media," *IEEE Trans. Multimedia*, vol. 8, no. 2, pp. 390-404, 2005.

[15] F. Fu and M. van der Schaar, "Decomposition Principles and Online Learning in Cross-Layer Optimization for Delay-Sensitive Applications", *IEEE Trans. Signal Process.*, to appear.

[16] Dejan V. Djonin, Vikram Krishnamurthy, "Q-learning algorithms for constrained Markov decision processes with randomized monotone policies: application to MIMO transmission control," *IEEE Transactions on Signal Processing* 55(5-2): 2170-2181 (2007)

[17] N. Salodkar, A. Bhorkar, A. Karandikar, and V. S. Borkar, "An on-line learning algorithm for energy efficient delay

constrained scheduling over a fading channel," *IEEE Journal on Selected Areas in Communications* 26(4): 732-742 (2008)

[18] E. Altman, "Constrained Markov decision processes: stochastic modeling," London: Chapman and Hall, CRC, 1999.

[19] D. P. Bertsekas, "Dynamic programming and optimal control," 3$^{rd}$, Athena Scientific, Massachusetts, 2005.

[20] Q. Zhang, S. A. Kassam, "Finite-state Markov Model for Reyleigh fading channels," *IEEE Trans. Commun.* vol. 47, no. 11, Nov. 1999.

[21] D. Bertsekas, and R. Gallager, "Data networks," Prentice Hall, Inc., Upper Saddle River, NJ, 1987.

[22] D. M. Topkis, "Supermodularity and complementarity," Princeton University Press, Princeton, NJ, 1998.

[23] R. S. Sutton, and A. G. Barto, "Reinforcement learning: an introduction," Cambridge, MA:MIT press, 1998.

[24] V. S. Borkar, S. P. Meyn, "The ODE method for convergence of stochastic approximation and reinforcement learning," *SIAM J. Control Optim*, vol 38, pp. 447-469, 1999.

[25] A. Siem, D. Hertog and A. Hoffmann, "A method for approximating univariate convex functions using only function value evaluations, ". Online, available at SSRN: http://ssrn.com/abstract=1012289.

[26] A. Siem, D. Hertog and A. Hoffmann, "Multivairate convex approximation and least-norm convex data-smoothing," *Lecture Notes in Computer Science*, vol. 3982, pp. 812-821, 2006.

[27] W. Powell, A. Ruszcznski, and H. Topaloglu, "Learning algorithms for separable approximation of discrete stochastic optimization problems," *Mathematics of Operations Research*, vol. 29, no. 4, pp. 814-836, Nov. 2004.

[28] Simao, H. P. and W. B. Powell, "Approximate dynamic programming for management of high value spare parts", *Journal of Manufacturing Technology Management,* Vol. 20, No. 9 (2009).

[29] F. Fu and M. van der Schaar, "Structural solutions for cross-layer optimization of wireless multimedia transmission," *Tech-Report*, available on: http://medianetlab.ee.ucla.edu/UCLATechReport_CLO.pdf.

[30] F. Fu, D. Turaga, O. Verscheure, M. van der Schaar, and L. Amini, "Configuring Competing Classifier Chains in Distributed Stream Mining Systems" *IEEE Journal of Selected Topics in Signal Process. (JSTSP)* , vol. 1, no. 4, pp. 548-563, Dec. 2007.

[31] N. Mastronarde and M. van der Schaar, "A Queuing-Theoretic Approach to Task Scheduling and Processor Selection for Video Decoding Applications," *IEEE Trans. Multimedia*, vol. 9, no. 7, pp. 1493-1507, Nov. 2007.

[32] Rahul Urgaonkar, Ulas C. Kozat, Ken Igarashi, Michael J. Neely, "Dynamic Resource Allocation and Power Management in Virtualized Data Centers," to appear in *Proc. of IEEE/IFIP NOMS 2010* , Osaka, Japan, April 2010.

[33] F. Fu and M. van der Schaar, "A Systematic Framework for Dynamically Optimizing Multi-User Video Transmission," *IEEE J. Sel. Areas Commun.*, to appear.

[34] B. George and J. Gwilym "Time series analysis: forecasting and control,"San Francisco: Holden-Day, 1970.

[35] M. Grant and S. Boyd. CVX: Matlab software for disciplined convex programming (web page and software). http://stanford.edu/~boyd/cvx, June 2009.

[36] Kleinrock, L., "Queueing Systems, Volume I: Theory," Wiley Interscience, New York, 1975.